\documentclass[conference]{IEEEtran}

\usepackage{amsmath}
\usepackage{algorithmic}
\usepackage{array}
\usepackage{amsfonts}

\usepackage{tikz}
\usepackage{url}

\begin{document}

\title{Deep Reinforcement Learning based Autonomous Decision-Making for Cooperative UAVs: A Search and Rescue Real World Application}

\author{\IEEEauthorblockN{Thomas Hickling, Maxwell Hogan, Abdulla Tammam, and Nabil Aouf}
\IEEEauthorblockA{School of Science and Technology\\
City St George's University of London\\
London, United Kingdom\\
Email: Tom.Hickling, Maxwell.Hogan, Abdulla.Tammam, Nabil.Aouf@city.ac.uk}
}

\maketitle

\begin{abstract}
This paper presents the first end‑to‑end framework that combines guidance, navigation, and centralised task allocation for multiple UAVs performing autonomous search‑and‑rescue (SAR) in GNSS‑denied indoor environments. A Twin Delayed Deep Deterministic Policy Gradient controller is trained with an Artificial Potential Field (APF) reward that blends attractive and repulsive potentials with continuous control, accelerating convergence and yielding smoother, safer trajectories than distance‑only baselines. Collaborative mission assignment is solved by a deep Graph Attention Network that, at each decision step, reasons over the drone–task graph to produce near‑optimal allocations with negligible on‑board compute. To arrest the notorious Z‑drift of indoor LiDAR‑SLAM, we fuse depth‑camera altimetry with IMU vertical velocity in a lightweight complementary filter, giving centimetre‑level altitude stability without external beacons. The resulting system was deployed on two 1~m‑class quad‑rotors and flight‑tested in a cluttered, multi‑level disaster mock‑up designed for the NATO‐Sapience Autonomous Cooperative Drone Competition. Compared with prior DRL guidance that remains largely in simulation, our framework demonstrates an ability to navigate complex indoor environments, securing first place in the 2024 event. These results demonstrate that APF‑shaped DRL and GAT‑driven cooperation can translate to reliable real‑world SAR operations.


\end{abstract}

\begin{figure}[t]
    \centering
    \includegraphics[width=1.0\linewidth]{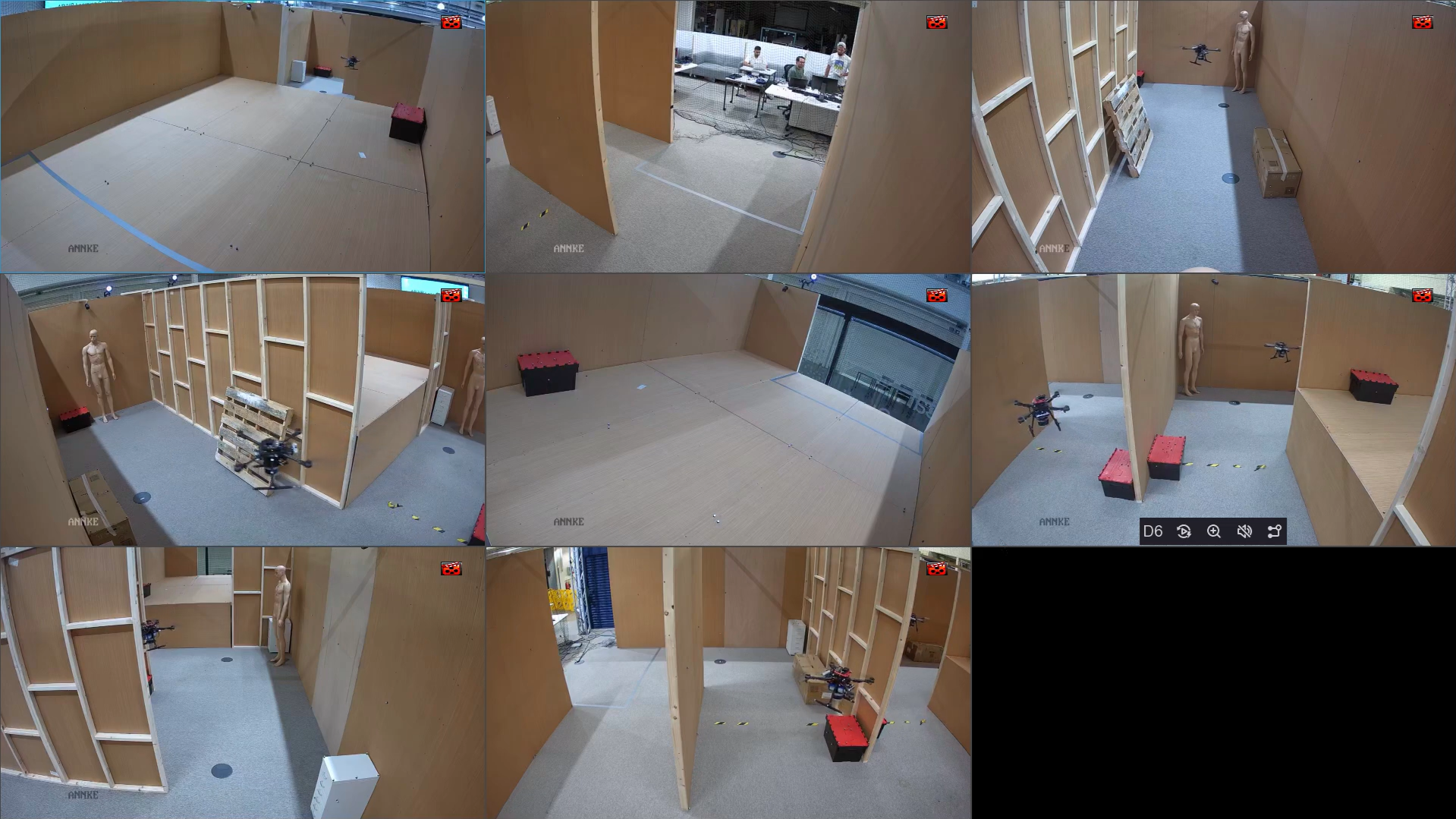}
    \caption{The arena built for the Sapience competition as seen by the eight cameras used for monitoring the UAVs}
    \label{fig:arena}
\end{figure}

\begin{figure}[t]
    \centering
    \includegraphics[width=0.75\linewidth]{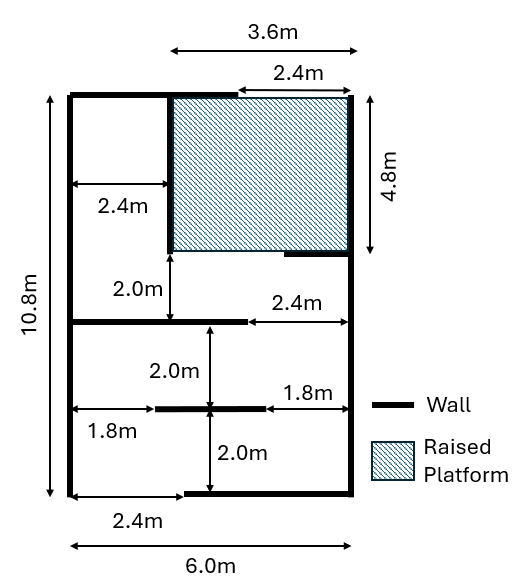}
    \caption{The floor plan for the constructed building}
    \label{fig:layout}
\end{figure}

\section{Introduction}
The adoption of Uncrewed Aerial Vehicles (UAVs) has grown in critical applications such as search and rescue (SAR) operations due to their ability to access hard-to-reach areas and quickly provide real-time situational awareness, \cite{scientific_american_drones}. However, operating in environments denied by the Global Navigation Satellite System (GNSS), such as indoor settings, presents unique challenges for autonomous UAVs, which require the development of intelligent and robust navigation and task management strategies. In recent years, Deep Reinforcement Learning (DRL), has emerged as a promising approach to handling complex decision-making tasks in such scenarios, particularly for cooperative UAV systems \cite{electronics10090999}.

This paper presents the development and deployment of a cooperative drone system that adopts deep learning strategies to achieve autonomous navigation, guidance, and planning. This development is conducted within the NATO-funded Sapience program through an Autonomous Cooperative Drone Competition, \cite{sapience}, which promotes technological advancements as part of NATO’s humanitarian goals. The competition focuses on improving cooperative autonomous UAV capabilities in GNSS-denied environments for search and rescue applications. Teams were challenged with tasks such as mapping, object/person detection and localisation, and the delivery of medical aid. These tasks were to show the benefit of using autonomous systems to perform tasks such as medical aid delivery without subjecting operators to unnecessary risk. The arena created for these tasks can be seen in Fig. \ref{fig:arena} with its layout being shown in Fig. \ref{fig:layout}

The system developed for this competition employs two UAVs, each equipped with 13-inch propellers, two Intel RealSense D455f cameras \cite{IntelRealSenseD455f}, a Velodyne 16-line puck Light Detection and Ranging (LIDAR) sensor \cite{velodyne_vlp16_manual}, and an NVIDIA Jetson Orin Nano \cite{NVIDIAJetsonOrinNano2024}. The DRL-based guidance for each UAV was trained in a simulation environment using the Twin Delayed Deep Deterministic Policy Gradient (TD3) algorithm \cite{fujimoto2018addressing} with AirSim \cite{airsim2017fsr}, a widely used UAV simulation platform.

A vital component of a cooperative drone system is how the agents cooperate. This paper presents a DRL-trained graph-based task allocation mechanism that optimises real-time coordination between multiple agents. This solution allows the two drones to effectively divide search areas and dynamically adjust their strategies based on environmental feedback, leading to efficient and timely exploration in SAR scenarios. The task allocator is trained alongside the DRL guidance model, enabling the system to adapt to various indoor challenges and obstacles.

A form of odometry is needed to fly in an indoor GNSS-denied environment. Due to the constructed environment, the traditional Visual Simultaneous and Mapping (VSLAM) was unsuitable due to a lack of features for mapping, \cite{tourani2022vslam}. This led to LIDAR being explored in the form of LIDAR Simultaneous and Mapping (LIDAR-SLAM), \cite{grisetti2005improving} and then adapted to deal with narrow corridors, which cause drifting in the vertical coordinates, to allow accurate positioning.

This paper aims to contribute to the development of autonomous drone systems for search and rescue operations in environments denied by GNSS by developing and deploying cooperative drones that integrate advanced navigation, mapping, guidance, and task allocation solutions. The presented system combines LIDAR-SLAM for robust localisation and mapping, DRL-trained guidance for autonomous navigation, and a Graph Attention Network (GAT)-based task allocator for efficient multi-UAV coordination. By demonstrating these capabilities in the context of the NATO SAPIENCE Autonomous Cooperative Drone Competition, this work highlights the potential of UAV technologies to revolutionise search and rescue missions, addressing critical challenges such as agent cooperation, real-time navigation, obstacle avoidance, and efficient resource allocation, while aiming to improve response times, operational efficiency, and safety in life-saving scenarios.

\paragraph*{Key Contributions.}
Unlike prior work that validates DRL controllers only in small-scale labs or simulations, we report the end-to-end real-world deployment of a cooperative DRL policy for autonomous SAR with multiple UAVs.  Our main novelties are:

\begin{enumerate}
    \item \textbf{Extensive real‑world evaluation}.  We flew two medium-sized (1.0m at widest) quad‑rotors across a large, structurally complex disaster mock‑up. This paper details the lessons learned from tackling the sim-2-real problem in multi-drone guidance.

    \color{black}
    \item \textbf{APF-referenced action-alignment reward for continuous velocity-level UAV control.} We propose an APF-referenced shaping objective in which an APF-inspired scheme produces a step-wise reference action \(a_t^\star\) in the same control space as the UAV commands, and the DRL agent is rewarded for aligning its executed action \(a_t\) with this reference. The design includes obstacle-proximity mode switching (attractive vs.\ repulsive behaviour), reward clipping for training stability, and terminal overrides for goal completion and collision. We demonstrate that this reward enables stable training and real-world deployment of a continuous-velocity guidance policy in a cluttered indoor environment where a standard distance-to-goal reward did not produce a deployable controller under the same training constraints.
    \color{black}

    \item \textbf{Centralised task allocation via Graph‑Attention DRL}.  We formulate the cooperative search as a one‑hop message‑passing problem and train GAT‑actor–critic that centrally assigns tasks while each UAV executes local low‑level policies.

    \item \textbf{Altitude‑drift suppression for indoor LiDAR‑SLAM}.  To mitigate the well‑known Z‑drift of 3‑D LiDAR SLAM indoors, we fuse a depth camera–derived altitude with the IMU’s vertical velocity via a lightweight complementary filter, recovering a drift‑free global $z$ estimate.
\end{enumerate}

Taken together, these novelties enable, for the first time, robust, collaborative DRL‑guided SAR flight in complex, GPS‑denied indoor environments.

\section{Related Work}
\subsection{Deep Reinforcement Learning for Autonomous Guidance and Obstacle Avoidance}
UAV guidance using DRL has been studied,  \cite{azar2021drone} and the developed approaches generally fall into two categories: discrete action spaces \cite{zhu2024uav}, where the UAV’s actions are restricted to specific movements (e.g., move forward, move right), and continuous action spaces \cite{hu2022obstacle}, where the UAV can select from a range of values for each action, allowing more nuanced control. The choice of DRL algorithms depends on the type of control required. For discrete action spaces, algorithms such as Deep Q-Networks (DQN), \cite{mnih2015human}, \cite{zhu2024uav} are suitable, while for continuous action spaces, algorithms like Deep Deterministic Policy Gradient (DDPG) \cite{lillicrap2015continuous}, \cite{hickling2023robust}, Proximal Policy Optimisation (PPO) \cite{schulman2017proximal}, \cite{hu2022obstacle}, and Twin Delayed Deep Deterministic Policy Gradient (TD3) \cite{fujimoto2018addressing}, \cite{he2021explainable} are more appropriate.

DRL typically employs unsupervised learning, relying solely on states, actions, and rewards observed during training. This unsupervised approach allows DRL to produce novel solutions unbounded by existing methods; however, it often requires significantly more samples to train effectively, leading to extended training times, especially in complex environments like for UAV guidance.

To address this inefficiency, researchers have incorporated expert knowledge into DRL training, enhancing performance through pre-training. Techniques such as Deep Q Learning from Demonstrations (DQfD) \cite{hester2018deep} and Deep Deterministic Policy Gradients from Demonstrations (DDPGfD) \cite{vecerik2017leveraging} leverage expert-labelled data to pre-train networks, reducing reliance on extensive reward engineering while preserving the benefits of DRL.

\color{black}
Artificial Potential Fields (APFs) are widely used in UAV navigation as structured priors for goal-seeking and obstacle avoidance. When combined with DRL, the literature largely falls into three recurrent integration patterns: (i) using APF to guide the policy during training, (ii) embedding APF quantities within the reward as a shaping signal, and (iii) treating APF as an auxiliary controller or planner that is fused with, or adapted by, a learned policy.

One pragmatic approach is to generate APF "non-expert" actions that steer learning early on, whilst gradually reducing this influence to restore exploration. NPE-DRL \cite{zhang2024npe} follows this philosophy in a discrete action space, whereas DDPG-APF \cite{hickling2023robust} demonstrates an analogous idea under continuous control. In both cases, the APF acts as a curriculum-like scaffold: it provides a stabilising directional cue in the initial phases of training, and is annealed so that the final policy is not constrained by the hand-crafted field. Empirically, these approaches typically improve sample efficiency and reduce the frequency of catastrophic exploration during early training.

A second class of methods incorporates APF structure directly into the reward function, using potential-inspired terms to densify feedback beyond sparse goal completion rewards. Li and Wu \cite{li2020path} integrate APF-inspired obstacle avoidance with a line-of-sight tracking objective, combining task-driven terms with safety-oriented penalties and action costs. Dong et al. \cite{dong2023uav} similarly decompose reward into attractive, repulsive, and directional alignment components, thereby encouraging both progress and locally sensible avoidance. These formulations can be effective in guiding the policy towards safer behaviour, but they remain sensitive to the relative weighting of reward components and, in several instances, are paired with discretised manoeuvre sets (e.g., a finite set of yaw-rate and vertical velocity commands), which can limit fine control authority in confined environments.

A third integration pattern uses APF to bias or constrain actions at execution time. Shield-DDPG, for example, augments the action generation process with an APF-inspired attractive component and employs a safety ``shield'' that overrides unsafe actions \cite{li2023uav}. Such designs can provide explicit constraint satisfaction, although the deployed behaviour is partly determined by the override logic and may differ from the learned policy. Related hybrid approaches treat APF as a planner whose behaviour is adapted by learning. PG-ITD3 combines a TD3 Variant with potential-field guidance and introduces mechanisms to mitigate classic APF failure modes (e.g., local minima and goal non-reachability), while also accounting for kinematic constraints \cite{ren2025pg}. In these frameworks, APF provides a strong inductive bias for navigation, while learning adjusts parameters or higher-level decisions to improve robustness.

This paper is most closely related to APF-guided and APF-shaped DRL methods, but differs in where the potential field is placed in the learning loop and in the targeted deployment setting. In contrast to approaches that treat APF as an external "teacher" through action injection (e.g., NPE-DRL \cite{zhang2024npe} and DDPG-APF \cite{hickling2023robust}), we integrate APF directly into the guidance objective by shaping reward in the same continuous action space used for deployment. Relative to reward-only potential shaping, we emphasise that our formulation produces dense, physically meaningful feedback aligned with the commanded control variables, which is particularly important for indoor, GNSS-denied flight, where tight clearances and non-ideal dynamics demand smooth and responsive velocity and yaw-rate control. Finally, unlike safety-override schemes, our intention is for the learned policy to internalise the desired avoidance behaviour rather than routinely relying on an external action filter, improving consistency between training and fielded execution. We provide real-world experimental data to back up the 

\color{black}

Talha et al.\ \cite{10.1007/978-3-031-22695-3_51} used DRL to generate waypoints for a UAV in an AirSim SAR simulation, combining learned guidance with classical planners and detectors. In contrast, our controller outputs direct velocities and is validated in a cluttered indoor setting where fine control is essential, rather than the comparatively forgiving outdoor scene they considered.

A persistent weakness in the literature is the scarcity of real-world indoor tests. Phueakthong et al.\ \cite{10949537} opted for a position-based (2‑D) controller to limit aggressive manoeuvres, feeding a 26‑zone LiDAR scan and target pose into an FCN. Their platform (1.9\,kg, 0.1\,m/s) successfully rounded a corner, but the action/state spaces were deliberately constrained. We expand both: yaw and $z$ velocity are controlled, a depth image and a denser LiDAR scan are processed, and our 4\,kg UAVs fly at up to 1.0\,m/s in a far more complex layout.

Other efforts have begun to probe sim‑to‑real transfer. Using PPO, \cite{10553074} trained a micro‑drone to orbit a column and injected Gaussian noise to harden the policy against sensor errors, improving real-world transfer in a small arena. While valuable, that study still operates at a miniature scale and with a limited task.

Real-world DRL guidance is more common in outdoor settings. He et al.\ \cite{he2021explainable} emphasise explainability and demonstrate tree‑traversal with a 2‑D flight envelope. Their contribution complements ours—interpretability versus operational breadth—but does not align with the scope of our indoor, multi-UAV deployment.

In summary, prior work either remains in simulation, restricts dynamics/action spaces, or tests at a small scale. Our study addresses this gap with full‑scale, multi‑agent DRL guidance in a GNSS-denied indoor environment, extensive real-world trials, and centralised GAT-based task allocation.

\subsection{Deep Reinforcement Learning for Graph-Based Task Allocation}
The use of multiple UAVs in cooperative missions has been explored to enhance task efficiency and flexibility \cite{sargolzaei2020control}. Early task allocation methods relied on rule-based or heuristic optimisation approaches, including centralised systems where a single controller assigns tasks to each UAV and decentralised systems where UAVs independently select their tasks based on predefined rules \cite{zhao2015heuristic}. While these methods provided foundational insights, they often faced scalability challenges and lacked the adaptability needed for dynamic environments \cite{cui2019multi}. As a result, researchers have increasingly turned to Machine Learning (ML) approaches, particularly DRL \cite{mao2024dl}, to address these limitations by enabling UAVs to learn adaptive task allocation policies that are robust to environmental changes.

Within DRL-based task allocation, various algorithms have been developed for multi-UAV coordination, including Markov Decision Process \cite{mao2024dl}, and Multi-Agent Deep Deterministic Policy Gradient \cite{lowe2017multi}. These DRL approaches rely on reward structures designed to optimise objectives such as maximising task coverage, minimising collisions, and improving efficiency in multi-agent settings. The adaptability of DRL allows UAVs to dynamically allocate tasks in response to changes in mission objectives or environmental constraints, providing advantages over classical optimisation techniques in complex, multi-agent scenarios \cite{mao2024dl}.

Alshaboti and Baroudi\cite{alshaboti2021multi} compare two distributed multi-robot task-allocation (MRTA) strategies—a fuzzy-inference auction and an adaptive multi-threshold rule—in dynamic, single-task/single-robot settings where tasks arrive online and multiple objectives (distance, load balance, “quality” match) must be traded off. Their Mamdani FIS compresses distance, load, and quality-gap metrics into a scalar bid, while the threshold scheme allows each robot to decide locally. Simulations with Khepera III robots in Webots show that the threshold method cuts the travelled distance by roughly 10\% but suffers from slightly worse load balance, whereas the auction achieves tighter balancing with similar quality of satisfaction. Both approaches, however, presuppose reliable communication, relatively simple robots, and GPS/landmark localisation, and they do not model rich perception or high-dimensional action spaces. In contrast, our work allocates tasks centrally via a graph-attention actor–critic that reasons over dense edge costs and heterogeneous node states, and—critically—validates the policy on full-scale UAVs in a GNSS-denied indoor environment rather than in simulation. This highlights a complementary trajectory: moving from scalar bids and hand-crafted fuzzy rules toward learned, end-to-end differentiable allocators embedded in a real SAR pipeline.

Graph-based representations have become a key approach in modelling spatial layouts and task relationships within multi-UAV systems \cite{GNN-EnhancedDRL}. Graphs provide a natural way to represent complex spatial relationships, with nodes representing specific locations, tasks, or UAVs, and edges representing viable paths or task dependencies. This structure is particularly beneficial in environments with intricate layouts, dynamic obstacles, or multiple task dependencies, as it allows for more informed decision-making based on the UAVs’ spatial context and objectives.

Graph Attention Networks (GATs) \cite{velickovic2017graph}, \cite{GA-DRL} improve upon traditional Graph Convolutional Networks (GCNs) by incorporating an attention mechanism that allows the model to assign varying levels of importance to different nodes and edges. This is especially useful in UAV task allocation, where certain nodes (tasks) or edges (paths) may be more critical than others for efficient navigation and task completion. By selectively focusing on the most relevant nodes, GATs enable prioritisation of tasks in real-time, enhancing coordination and efficiency in multi-agent systems \cite{shao2021graph}.

Ryu et al.\cite{ryu2020multi} propose HAMA, a multi‑agent actor–critic architecture that stacks hierarchical graph attention layers—separate inter‑agent and inter‑group attentions—to encode relationships and enable scalable, transferable policies in mixed cooperative–competitive games. They train under CTDE and show superior performance and transfer across predator–prey benchmarks (3v1, 3v3, “more‑the‑stronger”) compared to MADDPG, MAAC, and heuristics, while also interpreting strategies via learned attention weights. In contrast, our allocator is centralised (single forward pass assigns tasks to all UAVs) and is validated on full-scale robots in a GNSS-denied indoor SAR scenario, addressing sim-to-real, rich perception, and safety constraints that are absent from purely simulated domains.

The combination of GATs and DRL leverages the strengths of both approaches: GATs provide the structural awareness necessary for understanding task relationships, while DRL offers a flexible framework for learning task allocation policies based on trial-and-error feedback. This paper will explore how to combine these methods to develop a task allocator that enables UAV cooperation in indoor environments.

\subsection{Adapted LIDAR-SLAM for Navigation}
The "corridor problem" in LIDAR-SLAM is a known issue that occurs when using LIDAR for SLAM in narrow, featureless corridors. The problem arises because the lack of unique, distinguishable features in a long, narrow corridor can make it difficult for the LIDAR-SLAM system to localise and distinguish between different parts of the environment accurately.

Zhang et al. \cite{radarlidar} solved this issue by utilising radar to create more features with different materials and shapes, returning different radar returns. These additional features help mitigate common issues associated with the corridor problem. By leveraging different feature extraction methods, the limitations of one approach can be compensated by the other. This paper proposes a solution to Z-drift caused by the corridor problem through sensor fusion and converting relative positioning into the global positioning system used by LiDAR-SLAM mapping.

\section{Autonomous Guidance in UAVs}

Autonomous guidance for UAVs is traditionally facilitated by algorithms such as Dijkstra, A*, and APF. Recent advances in ML, particularly DRL, have enhanced UAVs' capabilities to make complex real-time decisions in search and rescue, mapping, and inspection.

\subsection{Deep Reinforcement Learning}

DRL combines reinforcement learning (RL) principles \cite{sutton2018reinforcement} with Deep Neural Networks (DNNs) to approximate policies or value functions, enabling agents to operate in large state-action spaces. In DRL, an agent interacts with an environment, selects actions, and receives feedback in the form of rewards, learning an optimal policy to maximise cumulative reward. This approach is particularly valuable in complex environments unsuitable for traditional tabular methods as the large state-action spaces take unfeasible amounts of computer memory.

Key components of DRL include:
\begin{itemize}
    \item \textbf{Agent:} The decision-maker in the environment.
    \item \textbf{Environment:} The external system with which the agent interacts.
    \item \textbf{State, Action, Reward:} Indicators of the environment’s current status, possible moves, and feedback for success.
    \item \textbf{Policy:} The strategy the agent uses to decide actions.
\end{itemize}

Popular DRL algorithms, such as DQN \cite{mnih2015human}, PPO \cite{schulman2017proximal}, and TD3 \cite{fujimoto2018addressing}, each address issues related to stability, exploration, and sample efficiency.





\subsubsection{Twin Delayed Deep Deterministic Policy Gradient}
TD3\cite{fujimoto2018addressing} refines DDPG\cite{lillicrap2015continuous} for continuous control by (i) clipped double Q‑learning, which updates from the smaller of two critic estimates to curb overestimation, (ii) delayed actor updates that let critic values stabilise, and (iii) target‑policy smoothing, adding small noise to target actions to regularise Q targets.  The agent stores transition tuples in a replay buffer, updates both critics and the actor using mini-batches in an off-policy loop, and outputs smooth, high-precision control signals, making TD3 well-suited to the cooperative UAV guidance task studied here.

\subsubsection{Artificial Potential Field Based Reward Structure}

The design of the reward system is vital to successfully training a DNN using DRL techniques, \cite{ng1999policy}. Small changes in how the reward is allocated can significantly change the training result. In DRL training, rewards are typically shaped by specific actions and states. Reward shaping is complicated when applied to training for complex goals and environments. Therefore, looking for novel reward-shaping methods to maximise DRL's promise is crucial.

The Artificial Potential Field (APF) \cite{khatib1986real} is a well-known technique used in robotics for path planning and obstacle avoidance. In the APF approach, the environment is modelled as a potential field where attractive and repulsive forces guide the agent (in this case, a drone) towards the goal while avoiding obstacles. The goal generates the attractive potential, which pulls the agent towards it, while the repulsive potential is generated by obstacles pushing the agent away. 

The potential field is defined as a scalar function over the space, and the agent moves in the direction that minimises this potential. Mathematically, the total potential \( U \) is typically composed of an attractive potential \( U_{\text{att}} \) and a repulsive potential \( U_{\text{rep}} \):

\begin{equation}
U(q) = U_{\text{att}}(q) + U_{\text{rep}}(q),
\end{equation}
where \( q \) represents the position of the drone. The attractive potential is often designed to decrease as the drone approaches the goal, and the repulsive potential increases as the drone approaches obstacles. 

The \textbf{resultant force} acting on the drone is derived from the gradient of the total potential function:

\begin{equation}
F(q) = -\nabla U(q),
\end{equation}
which gives the direction of movement for the drone. This force represents the \textbf{optimal movement direction}, balancing attraction to the goal and repulsion from obstacles.

In a RL context, we can leverage the optimal movement provided by the APF to design an effective reward function for training a drone in a DRL framework. Since the APF provides a clear direction for the drone to move in, we can compare the action chosen by the drone to this optimal movement and use the similarity between the two as a measure of performance.

\subsection{Advantages of Using APF in Reward Function}

This approach offers several advantages:
\begin{itemize}
    \item \textbf{Guided Learning:} By using the APF as a reference for optimal movement, the agent can learn more efficiently as it is guided towards the correct actions early in training.
    \item \textbf{Smooth Trajectories:} APF typically generates smooth, collision-free paths, ensuring the agent learns to navigate safely in complex environments.
    \item \textbf{Hybrid Approach:} Combining DRL with APF leverages the strengths of classical control (APF) and learning-based approaches (DRL), providing a balance between predefined optimal behaviour and the adaptability of learning.
\end{itemize}

Thus, using APF as a reference for movement and designing a reward function around the similarity of the agent's actions to this reference can accelerate the learning process and improve the overall drone guidance performance, especially in obstacle-rich environments.

\section{Task Allocation}
Task allocation in robotics refers to assigning tasks to individual robots in a multi-robot system to achieve collective goals efficiently, \cite{korsah2013comprehensive}. This is a crucial problem, particularly in complex environments, where robots must collaborate, explore, or cover areas while minimising resource usage and avoiding conflicts. 

\subsection{Deep Reinforcement Learning for Task Allocation}
Neural networks, particularly DRL and other ML techniques, have increasingly been applied to solve task allocation problems in multi-robot systems \cite{gronauer2022multi}. They can learn complex patterns and adapt to dynamic, uncertain environments, making them well-suited for real-time task allocation.

In DRL-based task allocation, the task assignment is framed as a decision-making problem where the agent (drone or central controller) learns to allocate tasks through trial and error. A neural network is trained to predict which task should be assigned to which drone to maximise the system's overall performance, such as reducing task completion time or minimising energy usage.

The state can include the positions and states of drones, the state of the tasks and their positions, and any other sensor data. The action is then the orders for each agent in the system to perform either for the next step or string a series of tasks to be done. The reward structure can be used to guide the network to optimise for tasks to be selected in a way that tasks are not repeated and that the shortest path is taken to do all tasks.

\subsection{Overview of Graph Convolutional Networks (GCNs) and Graph Attention Networks (GATs)}

\subsubsection{Graph Convolutional Networks (GCNs)}

Graph Convolutional Networks (GCNs) \cite{kipf2016semi} are a type of neural network specifically designed to operate on graph-structured data. GCNs extend the concept of convolution from regular grid data, such as images, to irregular graph data, allowing for the processing of data where relationships between entities are naturally represented as graphs. 

In GCNs, each node's representation is updated by aggregating information from its neighbouring nodes. The layer-wise propagation rule for a GCN can be written as:

\begin{equation}
H^{(l+1)} = \sigma \left( \tilde{D}^{-1/2} \tilde{A} \tilde{D}^{-1/2} H^{(l)} W^{(l)} \right),
\end{equation}

Where:
\begin{itemize}
    \item \( H^{(l)} \) is the matrix of node features at layer \( l \).
    \item \( \tilde{A} = A + I \) is the adjacency matrix with added self-loops, where \( A \) is the adjacency matrix of the graph, and \( I \) is the identity matrix.
    \item \( \tilde{D} \) is the diagonal degree matrix of \( \tilde{A} \).
    \item \( W^{(l)} \) is the trainable weight matrix at layer \( l \).
    \item \( \sigma \) is an activation function, such as ReLU.
\end{itemize}

The GCN learns to propagate and combine features from neighbouring nodes, effectively capturing the structural information of the graph. However, GCNs apply uniform weighting to the neighbouring nodes, treating all neighbours equally during aggregation. This limitation can reduce the model's ability to focus on the most important nodes in complex graphs, potentially leading to suboptimal performance in certain scenarios.

\subsubsection{Graph Attention Networks (GATs)}

Graph Attention Networks (GATs) \cite{velickovic2017graph} introduce an attention mechanism to address the limitations of GCNs by allowing the model to learn the relative importance of neighbouring nodes. Instead of treating all neighbours equally, GATs compute attention coefficients for each edge, enabling the network to focus on more relevant neighbours when aggregating information.

The attention mechanism in a GAT layer computes attention coefficients \( \alpha_{ij} \) between node \( i \) and its neighboring nodes \( j \) as follows:

\begin{equation}
\alpha_{ij} = \frac{\exp \left( \text{LeakyReLU} \left( \mathbf{a}^T \left[ W \mathbf{h}_i \| W \mathbf{h}_j \right] \right) \right)}{\sum_{k \in \mathcal{N}(i)} \exp \left( \text{LeakyReLU} \left( \mathbf{a}^T \left[ W \mathbf{h}_i \| W \mathbf{h}_k \right] \right) \right)},
\end{equation}

where:
\begin{itemize}
    \item \( \alpha_{ij} \) represents the normalised attention coefficient between nodes \( i \) and \( j \).
    \item \( \mathbf{h}_i \) and \( \mathbf{h}_j \) are the feature vectors of nodes \( i \) and \( j \).
    \item \( W \) is a shared weight matrix.
    \item \( \mathbf{a} \) is a learnable vector for computing attention.
    \item \( \| \) denotes concatenation.
    \item \( \mathcal{N}(i) \) represents the neighbourhood of node \( i \).
\end{itemize}

The resulting attention coefficients \( \alpha_{ij} \) are used to compute a weighted sum of the neighbours' features, allowing the model to focus more on important connections. This attention-based aggregation results in node representations that better capture the varying importance of neighbouring nodes. Unlike GCNs, which
rely on a fixed adjacency matrix for propagation, GATs
do not require a predefined structure and can operate on
dynamically changing graphs. This makes GATs more
versatile for real-time applications where the graph topology may evolve over time.

\section{Global Navigation Satellite System Denied Odometry}

GNSS-denied odometry estimates a vehicle’s position, velocity, and orientation in environments where GNSS signals are unavailable or unreliable, such as indoor, underground, or dense urban areas \cite{mohamed2019survey}. For UAVs, this capability is essential for autonomous navigation, enabling precise manoeuvring and obstacle avoidance in confined spaces. GNSS-denied odometry is particularly valuable in applications requiring real-time localisation and mapping, such as infrastructure inspection, search and rescue, and disaster response.

Several methods are commonly used for GNSS-denied odometry. Visual odometry (VO) \cite{sahiliVSLAM} relies on camera data to track motion by analysing changes in visual features between frames, though its accuracy can diminish in feature-poor or low-light areas. IMU-based odometry \cite{mourikis2007multi} uses accelerometers and gyroscopes to estimate motion through changes in acceleration and rotational velocity, though it accumulates error (drift) over time, making it more suited for short-term estimates and often requires sensor fusion for long-term accuracy.

\subsection{LIDAR-SLAM}

LIDAR-SLAM \cite{grisetti2005improving, magnusson2009three, segal2009generalized} addresses limitations in feature-poor or low-visibility environments by using laser sensors to build a 3D map while simultaneously localising the UAV. This process involves scanning the surroundings to generate a 3D point cloud, which is continuously updated through scan matching as the UAV moves. Loop closure, a crucial aspect of SLAM, helps correct accumulated drift by recognising previously mapped areas, thus improving the consistency of maps and localisation over time. Additionally, graph-based optimisation refines the map and position estimates, reducing errors.

\subsubsection{Challenges in Height (Altitude) Estimation in Enclosed Environments}

Despite its advantages, LIDAR-SLAM faces challenges in enclosed spaces. Limited fields of view can reduce the data quality in tight areas, while feature-poor settings (e.g., long corridors) make it difficult to differentiate between similar-looking sections, leading to localisation errors \cite{li2024laser}. Drift remains a problem without frequent loop closures, and reflective surfaces like glass or metal can distort LIDAR measurements. Operating close to walls or obstacles may also cause incomplete scans, highlighting the need for sensor fusion with IMUs or visual odometry to enhance mapping accuracy in complex environments.

\color{black}
\section{Problem Formulation}
\label{sec:problem_formulation}

We consider a cooperative UAV system composed of two learned decision-making modules: (i) a continuous-control guidance policy executed onboard each UAV for collision-aware navigation; and (ii) a centralised task allocation policy executed at the ground station for assigning tasks to multiple UAVs. The guidance policy handles short-horizon motion selection under partial observability, whilst the allocator operates at a lower frequency to coordinate high-level tasking.

\subsection{Guidance as a Partially Observable Markov Decision Process}
Each UAV navigation episode is modelled as a partially observable Markov decision process (POMDP). At time-step \(t\), the UAV receives a multi-modal observation
\begin{equation}
o_t = \big( I_t^{\text{depth}},\, \ell_t^{\text{LiDAR}},\, p_t \big),
\end{equation}
where \(I_t^{\text{depth}}\) is the forward-facing depth image, \(\ell_t^{\text{LiDAR}}\) is a 1-D LiDAR sweep, and \(p_t\) is a compact positional state containing goal-related quantities (e.g., range/bearing and altitude error) together with recent action history. The policy outputs a continuous bounded action
\begin{equation}
a_t = [v_{x,t},\, \omega_t,\, v_{z,t}] \in [-1,1]^3,
\end{equation}
corresponding to forward velocity, yaw-rate, and vertical velocity commands. The UAV then transitions to a new state via the environment dynamics and receives an immediate reward \(r_t\).

\paragraph{APF-derived optimal action.}
We compute an APF-inspired ``optimal'' reference action \(a_t^\star\) from goal geometry and obstacle proximity. When the UAV is beyond a \(1\,\mathrm{m}\) proximity threshold from obstacles, the reference is governed by attractive terms:
\begin{align}
v_{x,t}^\star &= \mathrm{clip}(\cos(\theta_t),\, 0.0,\, 1.0), \\
\omega_t^\star &= \mathrm{clip}\!\left(\theta_t \cdot \frac{3}{\pi},\, -1.0,\, 1.0\right), \\
v_{z,t}^\star &= \mathrm{clip}\!\left(2\,(z_{\text{goal}}-z_t),\, -1.0,\, 1.0\right),
\end{align}
where \(\theta_t\) is the heading error between the UAV and the target direction. 

When an obstacle is within \(1\,\mathrm{m}\), we switch to a repulsive mode. The forward reference velocity is set to move away from obstacles according to the obstacle bearing \(\theta_{\text{obs}}\):
\begin{equation}
v_{x,t}^\star = -0.5 \cos(\theta_{\text{obs}}),
\end{equation}
and the yaw-rate reference follows a piecewise turning function \(\omega^\star(\theta_{\text{obs}})\) that turns away from the obstacle depending on its relative angle. 

\paragraph{Reward.}
The guidance reward is designed to encourage the UAV to align its executed action with the APF reference action. The base reward is
\begin{equation}
r^{\text{base}}_t = 1 - \sum_{i\in\{x,\omega,z\}} \big(a_{t,i} - a^\star_{t,i}\big)^2,
\end{equation}
which is then clipped for stability:
\begin{equation}
r_t = \mathrm{clip}(r^{\text{base}}_t,\, -1.5,\, 1.0).
\end{equation}
In addition, terminal outcomes override the clipped shaping reward: reaching a goal yields \(r_t = R_{\text{goal}}=2\), whilst colliding with an obstacle yields \(r_t = R_{\text{collision}}=-2\). 

\paragraph{Objective and termination.}
The guidance policy \(\pi_\phi(a_t|o_t)\) is trained to maximise the expected discounted return
\begin{equation}
J(\pi_\phi) = \mathbb{E}\left[\sum_{t=0}^{T-1} \gamma^t r_t\right],
\end{equation}
where episodes terminate on goal completion, collision, or a fixed horizon \(T\).

\subsection{Task Allocation as a Centralised Graph Decision Process}

Task allocation is cast as a centralised RL problem operating at decision epochs \(t\) over a fully-connected graph \(G_t=(V_t,E_t)\) comprising UAV nodes and task nodes. Node features encode completion/occupancy indicators, and edge attributes encode collision-free shortest-path distances between entities. The allocator outputs an assignment matrix
\begin{equation}
a_t = [a_{nk}] \in \{0,1\}^{N\times(M+1)},
\end{equation}
where \(a_{nk}=1\) assigns UAV \(n\) to task \(k\) (or to an ``idle'' task \(k=0\)), subject to \(\sum_k a_{nk}=1\) for each UAV.

The allocation reward credits productive completion whilst penalising inefficiency and collisions:
\begin{equation}
r_t = \Delta C_t - \lambda_d \Delta D_t - \lambda_c C_t,
\end{equation}
where \(\Delta C_t\) is the number of newly completed tasks, \(\Delta D_t\) is the incremental distance flown above a baseline, and \(C_t\) indicates collision events. Episodes terminate when all tasks are complete or a fixed horizon is reached.

\color{black}

\section{Methodology}
In this work, key technologies such as LIDAR-SLAM for real-time mapping and localisation, a DRL-based guidance AI for autonomous navigation, and a Graph Attention Network (GAT)-based task allocator for multi-UAV coordination were leveraged to develop a cooperative drone system for indoor SAR operations. The SAPIENCE competition looks to improve technologies like those mentioned above to enable UAVs to operate autonomously, avoid obstacles, and collaborate effectively to cover search areas, ensuring efficient exploration and critical aid delivery in environments where human intervention may be hazardous or infeasible.

\subsection{UAV and Sensor Suite}
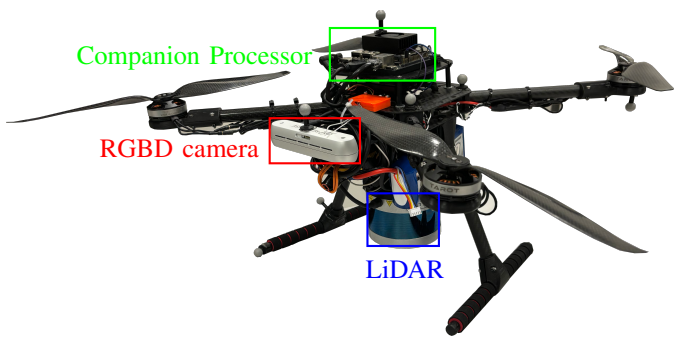
\begin{figure}[ht]
    \centering
    \begin{tikzpicture}
        \node[anchor=south west,inner sep=0] (image) at (0,0) {\includegraphics[width=1.0\linewidth]{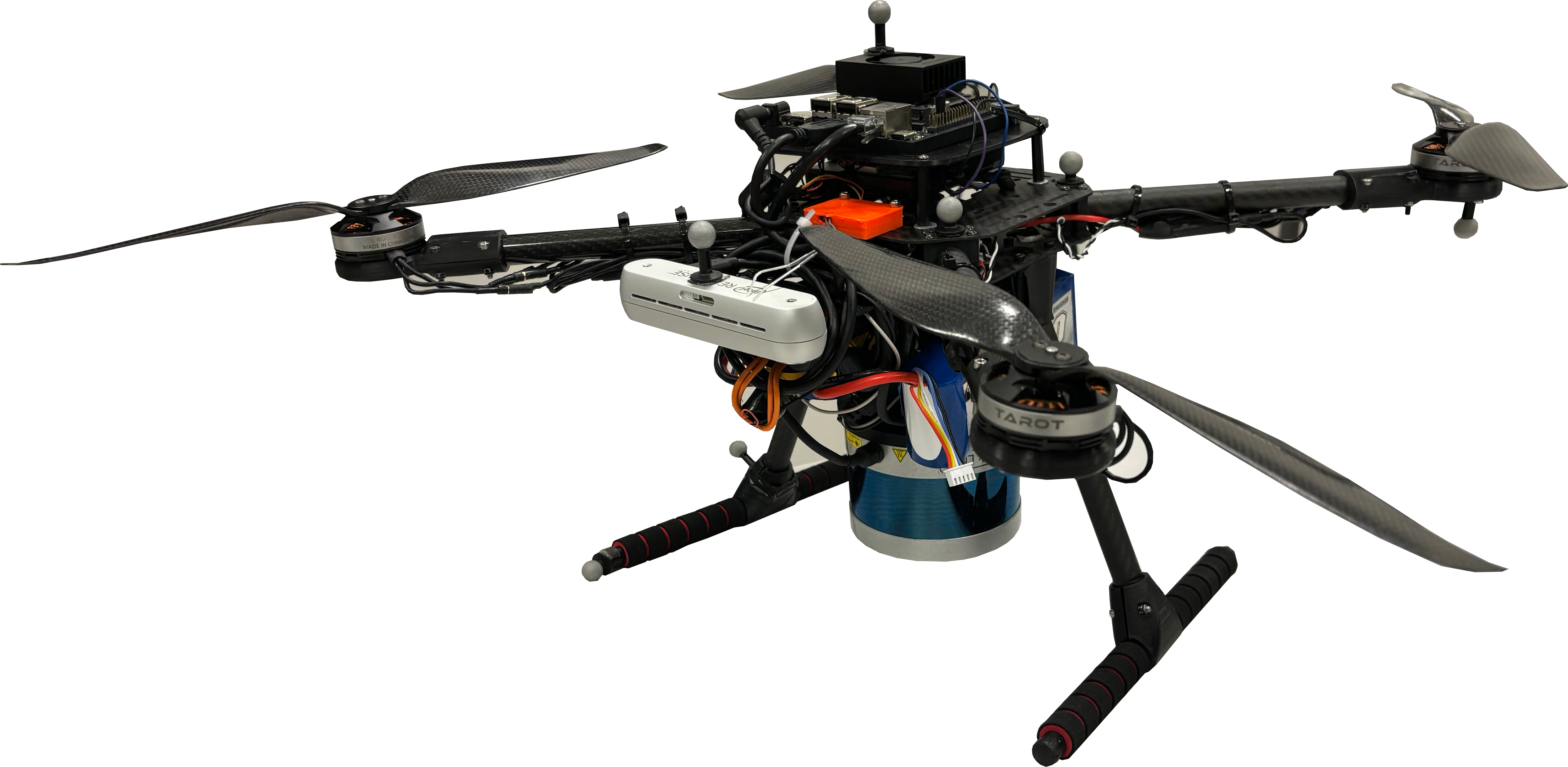}};
        
        \draw[red, thick] (3.5, 2.3) rectangle (4.7, 2.9); 
        
        \draw[blue, thick] (4.8, 1.2) rectangle (5.75, 1.9); 

        \draw[green, thick] (4.3, 3.4) rectangle (5.7, 4.1); 


        \node[text=red] at (2.3, 2.5) {RGBD camera};
        \node[text=blue] at (5.3, 0.9) {LiDAR};
        \node[text=green] at (2.5, 3.7) {Companion Processor};

    \end{tikzpicture}
    \caption{City University's Autonomous Drone.}
    \label{fig:Drone}
\end{figure}

Figure \ref{fig:Drone}, displays the drone developed for this work with annotations to show the additional equipment that was used to provide the platform with additional processing and sensing capabilities. The primary sensors include a Velodyne 16-line puck LiDAR \cite{velodyne_vlp16_manual} and an Intel RealSense D455f RGB-Depth camera \cite{IntelRealSenseD455f}. The companion processor is the NVIDIA Jetson Orin Nano \cite{NVIDIAJetsonOrinNano2024}, equipped with an NVIDIA Ampere GPU featuring 1024 CUDA cores and 32 Tensor Cores. This enables the processor to run advanced deep-learning algorithms and perform GPU-accelerated tasks efficiently. The systems of operation were developed using Isaac ROS \cite{NVIDIAIsaacROS2024}, which enabled the configuration of the sensors, interchange of commands to the flight controller, and embedding of the guidance and navigation algorithms developed in this work. In addition, this package also facilitated inter-process communication that accelerated prototyping and development without compromising real-time performance. 

Real-time positioning and mapping are made by integrating the lidarslam\_ros2 \cite{lidarslam_ros2} into the system. This provided the system with accurate positioning, including a heading, in order to undertake its assigned tasks. The difficulty in deploying the LiDAR-SLAM algorithm that is made available in this package is that it is susceptible to drift in the predicted height, given the absence of features. 

\subsection{LIDAR-SLAM Altitude Fix}
\color{black}
To fix this problem, the most obvious solution would be to use a laser range finder or a radar altimeter. We were unable to use these as a solution because the parts were not on the approved competition list, so another solution was needed. In the competition equipment list, a second Intel RealSense was allowed, so we moved on to using this as our range finder.

To utilise the RGB-D camera, we need to find a robust way to measure the correct distance. First, the median depth was calculated from five equidistant areas in the depth image. One of these was located in the centre of the image, and the other four were in the centre of each quadrant. The final depth was measured as the median of the five values and injected into the control loop. 

This method was used to improve the accuracy of the measured distance. The depth camera is a noisy sensor, with values that can fluctuate widely. So by using five areas and using the median values, we factor in outliers when taking a measurement. The camera's wider field of view also provides a clearer view of what is beneath the drone as it flies, compared to a more typical laser rangefinder. This means that when negotiating a ledge, we can be more certain when the edge is cleared.
\color{black}

The environment in which the drone was expected to operate was mostly flat, with multiple levels. To detect a change in level, a control loop was devised that would compare the vertical velocity measured by the flight controller's Inertial Measurement Unit (IMU) with the calculated velocity from concurrent depth measurements from the depth camera. Any significant deviations between these two values indicate a change in level, and the calculated relative depth is updated to reflect this.

\subsection{Guidance Deep Neural Network Architecture}
The actor network in our TD3 guidance scheme fuses three perception streams—depth images, LiDAR scans and positional/task variables—to output velocity commands for the UAV.

\subsubsection{Branch 1: Depth Images}
Depth frames are processed by eight 2‑D convolutional layers grouped into four blocks, each followed by max‑pooling. LeakyReLU activations and batch normalisation stabilise training \cite{lecun1998gradient,maas2013rectifier,ioffe2015batch}. The feature map is flattened and passed through two linear layers to yield a 128‑D vector, finished with a \texttt{tanh} activation.

\subsubsection{Branch 2: LiDAR Scan}
A 1‑D LiDAR sweep is handled by five 1‑D convolutional layers arranged in three blocks, again with LeakyReLU and batch normalisation. A final linear layer compresses the representation to 128 dimensions and applies \texttt{tanh}.

\subsubsection{Branch 3: Positional State}
A compact vector (goal range and bearing, altitude error, relative height, and previous $x$, $z$ and yaw actions) is mapped to a 128‑D embedding through a single fully connected layer with \texttt{tanh} activation.

\subsubsection{Fusion and Output}
The three 128‑D embeddings are concatenated (384‑D) and refined by five fully connected layers with LeakyReLU. A final linear layer with \texttt{tanh} outputs bounded $x$‑velocity, $z$‑velocity and yaw‑rate commands (-1 to 1), aiding stable control.

\subsubsection{Training Context}
This network serves as the actor (and target actor) in TD3 (Fig.~\ref{fig:td3}); the critics share the same architecture but concatenate the action at the fusion stage and regress a single Q‑value.

\subsubsection{Design Rationale}
\begin{figure}
    \centering
    \includegraphics[width=1.0\linewidth]{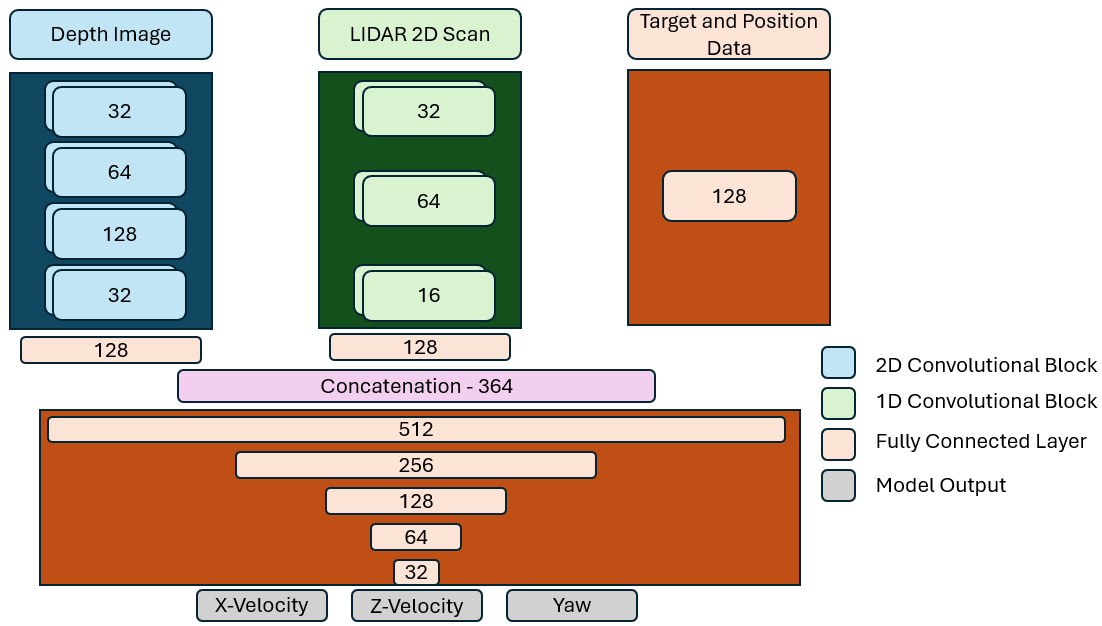}
    \caption{The architecture for the guidance AIs actor network}
    \label{fig:navmodel}
\end{figure}

\begin{figure}
    \centering
    \includegraphics[width=1.0\linewidth]{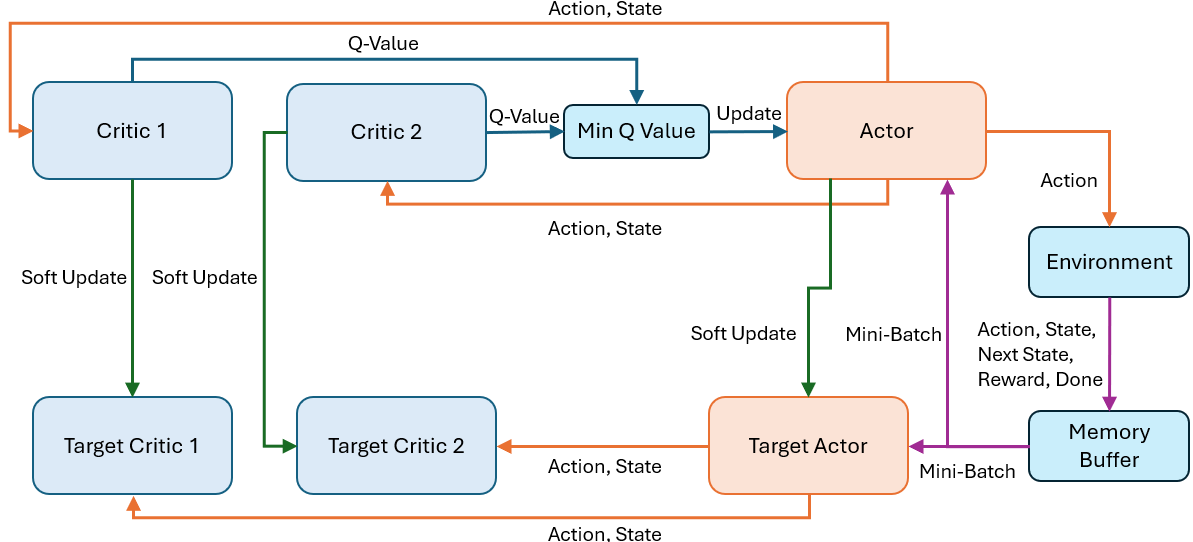}
    \caption{The TD3 Algorithm Architecture used for training the guidance AI}
    \label{fig:td3}
\end{figure}

Convolutions capture spatial structure in depth and LiDAR inputs, while batch normalisation and LeakyReLU mitigate internal covariate shift and vanishing gradients. The final \texttt{tanh} ensures smooth, bounded control signals, which is advantageous for precise indoor flight. The multi-branch design enables the exploitation of complementary sensing modalities without requiring hand-crafted feature engineering (Fig.~\ref{fig:navmodel}).

\subsection{Reward Function for the Guidance DRL}
The APF approach guides the UAV towards the target while avoiding obstacles by generating a set of forces that determine the desired actions for forward movement (\(x\)-velocity), turning (yaw speed), and altitude adjustment (\(z\)-velocity). The optimal actions are defined based on the distance to obstacles and the relative position of the target.

\subsubsection{Attractive Forces Towards the Target}

When the UAV is more than 1 meter away from any obstacles, the optimal action is determined by attractive forces that guide the UAV towards the target position. These forces are calculated as follows:

\begin{align}
\text{x}_{vel} &= \text{clip}(\cos(\theta), 0.0, 1.0), \\
\omega &= \text{clip}\left(\theta \cdot \frac{3}{\pi}, -1.0, 1.0\right), \\
\text{z}_{vel} &= \text{clip}\left(2 \cdot (\text{target}_z - \text{UAV}_z), -1.0, 1.0\right).
\end{align}

\noindent where:
\begin{itemize}
    \item \(\theta\) is the heading angle between the UAV and the target.
    \item \(\text{target}_z\) and \(\text{position}_z\) are the heights of the target and the UAV, respectively.
    \item \(\text{clip}(x, a, b)\) limits \(x\) to the range \([a, b]\).
\end{itemize}

The term \(\cos(\theta)\) ensures that the UAV moves faster when directly facing the target, with a movement restricted to forward directions due to the forward-facing depth camera. The yaw speed adjustment \(\text{force}_2\) scales the heading error, with the multiplier of 3 enhancing responsiveness, allowing for faster alignment with the target direction. The altitude adjustment \(\text{force}_3\) corrects the UAV's height relative to the target, adjusting at twice the rate of the height difference.

\subsubsection{Repulsive Forces for Obstacle Avoidance}

When the UAV is within 1 meter of an obstacle, the behaviour shifts to a repulsive mode, where forces are applied to move the UAV away from nearby obstacles and adjust its orientation. The repulsive forces are calculated based on the relative angle to the obstacle, \(\theta_{\text{obs}}\), which represents the bearing of the obstacle with respect to the UAV's forward direction.

The repulsive force in the forward direction (\(x\)-velocity) is calculated as:
\begin{equation}
\text{x}_{vel} = -0.5 \cdot \cos(\theta_{\text{obs}}),
\end{equation}
where \(\theta_{\text{obs}}\) is the angle to the closest detected obstacle. This formulation ensures that the UAV moves away from obstacles, with the negative sign indicating a backward movement. The cosine term reduces the backward speed when the obstacle is to the sides of the UAV, prioritising movement directly away from obstacles positioned directly in front.

The repulsive adjustment to the yaw speed, \(\omega(\theta_{\text{obs}})\), is designed to turn the UAV away from obstacles, depending on their relative position. It is defined as:
\begin{equation}
\omega(\theta_{\text{obs}}) =
\begin{cases}
0 & \text{if } \theta_{\text{obs}} = -\frac{\pi}{2}, \\
0.5 + \frac{\theta_{\text{obs}}}{\pi} & \text{if } -\frac{\pi}{2} < \theta_{\text{obs}} < 0, \\
-0.5 + \frac{\theta_{\text{obs}}}{\pi} & \text{if } 0 < \theta_{\text{obs}} < \frac{\pi}{2}, \\
0 & \text{if } \theta_{\text{obs}} = \frac{\pi}{2}.
\end{cases}
\label{eq:yaw_repulsive}
\end{equation}

This piecewise function ensures that the UAV adjusts its yaw speed based on the angle of the detected obstacle:
\begin{itemize}
    \item For obstacles directly to the left (\(\theta_{\text{obs}} = -\frac{\pi}{2}\)) or right (\(\theta_{\text{obs}} = \frac{\pi}{2}\)), no yaw adjustment is applied (\(\omega = 0\)).
    \item When the obstacle is slightly to the left (\(-\frac{\pi}{2} < \theta_{\text{obs}} < 0\)), a positive yaw adjustment is applied to turn the UAV right, away from the obstacle.
    \item For obstacles slightly to the right (\(0 < \theta_{\text{obs}} < \frac{\pi}{2}\)), a negative yaw adjustment is applied to turn the UAV left, away from the obstacle.
    \item The constants \(\pm 0.5\) provide a base turning speed, and the term \(\frac{\theta_{\text{obs}}}{\pi}\) adjusts the turning rate based on the angle, creating a smooth transition in yaw speed as the obstacle's relative position changes.
\end{itemize}

The repulsive force in the vertical direction (\(z\)-velocity) is calculated similarly to during the attraction periods of flight, shown in equation 9. These repulsive actions ensure the optimal action for the UAV to safely navigate away from obstacles is biased in the training. Combining backward movement with appropriate turning adjustments to maintain a safe distance from nearby objects.

\subsubsection{Calculating the reward} 
The reward function is designed to incentivise the UAV to follow the optimal actions closely while penalising collisions and rewarding goal achievement. The primary reward is based on the difference between the actual action of the UAV and the optimal action from the APF. The reward \(R\) is calculated as:

\begin{equation}
R = 1 - \sum (\text{action}_i - \text{action}_{\text{opt},i})^2
\end{equation}

where \(\text{action}_i\) represents the UAV’s action in the \(i\)-th dimension (e.g., \(x\)-velocity, yaw speed, and \(z\)-velocity), and \(\text{action}_{\text{opt},i}\) is the corresponding optimal action. Squaring the difference ensures all deviations are positive, with larger deviations penalised more severely. The reward drops off quickly as the UAV moves away from the optimal action, encouraging it to maintain alignment with the optimal path.

The reward is clipped to keep values between \(-1.5\) and \(1\), ensuring consistency with the bounds for task success and failure. Specifically, reaching a goal results in a reward of \(+2\), while colliding with an obstacle incurs a penalty of \(-2\). The final reward function is:

\begin{equation}
R = \text{clip}\left(1 - \sum (\text{action}_i - \text{action}_{\text{opt},i})^2, -1.5, 1\right),
\end{equation}

with additional rewards for task completion or penalties for collisions:
\[
R_{\text{goal}} = 2, \quad R_{\text{collision}} = -2.
\]

This reward structure encourages the UAV to take optimal actions, with significant deviations or collisions strongly penalised to guide effective learning. 

\subsection{Simulated Environment and Guidance Training}
The training of the proposed network architecture was conducted using a custom simulation environment developed in Epic's Unreal Engine 4 \cite{epicgames_unreal4}, integrated with the AirSim plugin developed by Microsoft. This simulation environment provides a realistic 3D setting for training the UAV using the TD3 algorithm which includes the perception branches shown in Fig. 2. The AirSim plugin enabled high-fidelity physics simulation and sensor emulation, allowing for realistic interactions between the UAV and the virtual environment.

The environment was designed with a variety of challenges to ensure robust learning. It featured a series of obstacles arranged along two different paths, with the path selection alternating throughout the training process to encourage exploration and adaptability. The obstacles included narrow corridors that required precise navigation, walls that needed to be circumnavigated, elevated sections that the UAV had to ascend, and barriers that required the UAV to manoeuvre over or under. This diversity of challenges aimed to expose the UAV to a wide range of scenarios that it might encounter in real-world applications.

The training process was conducted on a high-performance workstation equipped with an NVIDIA RTX A4500 GPU \cite{nvidia_rtx_a4500}, allowing for accelerated simulation and network optimisation. The training consisted of 1,000 episodes, during which the agent iteratively improved its guidance policy based on feedback from the environment. The reward system was designed to reinforce goal achievement and penalise collisions, facilitating the development of effective guidance strategies.

Upon completing the training, models that met a predefined threshold of successful goal completions were evaluated in a separate validation environment. This validation course was designed to test the generalisation capabilities of the trained models by presenting scenarios similar to, but distinct from, those encountered during training. Based on its ability to reach goals efficiently while avoiding obstacles, the model that demonstrated the highest performance in the validation environment was selected for deployment onto the physical UAV platform.

\begin{figure}
    \centering
    \includegraphics[width=0.8\linewidth]{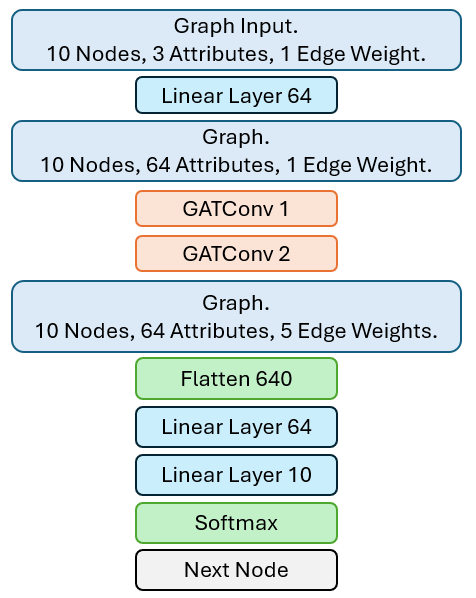}
    \caption{The network architecture of the GAT-Based model}
    \label{fig:GAT}
\end{figure}

\subsection{GAT‑based Task Allocation Architecture}
We cast cooperative task allocation as a single‑step, centralised RL problem trained with TD3. We adopted a central task allocator for two pragmatic reasons. \textit{(i)} The Sapience Round‑1 rules explicitly specified a centralised decision maker, so aligning with that requirement avoided penalties or re‑engineering late in the cycle. \textit{(ii)} Each UAV already maintains a single socket connection to the ground station for streaming sensor data for the mapping. Reusing that link to transmit task assignments was simpler and more reliable than implementing an additional peer‑to‑peer channel, reducing both software complexity and bandwidth overhead. The actor is a multi‑head Graph Attention Network (GATConv\cite{fey2019fast}); the two critics share the same graph encoder and concatenate the chosen action before the final linear layers.

\subsubsection{Problem Formulation}
At decision epoch $t$ the allocator observes a fully‑connected graph $\mathcal{G}_t=(\mathcal{V}_t,\mathcal{E}_t)$. The vertex set $\mathcal{V}_t=\mathcal{V}^{\textsc{uav}}_t\cup\mathcal{V}^{\textsc{task}}_t$ contains UAV nodes ($|\mathcal{V}^{\textsc{uav}}_t|=N$) and task nodes ($|\mathcal{V}^{\textsc{task}}_t|=M$). Each node carries a 3‑D feature vector $\,\mathbf{x}_v=[v_{\text{visited}},\,v_{\text{self}},\,v_{\text{peer}}]^\top$ indicating whether the task is completed and which UAV (if any) occupies it. Edge features are the shortest collision‑free distances $d_{ij}$.

\paragraph*{State}
The global state is the graph embedding:
$s_t=\langle\{\mathbf{x}_v\}_{v\in\mathcal{V}_t},\{d_{ij}\}_{(i,j)\in\mathcal{E}_t}\rangle$

\paragraph*{Action}
The allocator outputs an assignment matrix $\,a_t=\bigl[a_{nk}\bigr]\,\in\{0,1\}^{N\times(M+1)}$ where $a_{nk}=1$ assigns UAV $n$ to task $k$ (or to a dummy “idle’’ task $k=0$). Each UAV receives exactly one task: $\sum_{k}a_{nk}=1$.

\paragraph*{Reward}
Immediately after execution, we compute
\[
r_t = \underbrace{\Delta C_t}_{\text{tasks completed}}
       \;-\;\lambda_d\underbrace{\Delta D_t}_{\text{extra distance}}
       \;-\;\lambda_c\underbrace{C_t}_{\text{collision penalty}},
\]
where $\Delta C_t$ is the number of newly finished tasks, $\Delta D_t$ the incremental distance flown above a heuristic baseline, and $C_t$ a binary collision indicator. Positive reward is thus tied to productive, efficient, and safe allocations.

\paragraph*{Terminal condition}
Episodes terminate when all tasks are completed \emph{or} when a fixed horizon $T$ is reached.

\subsubsection{Network Details}
A preprocessing linear layer expands node features to 64 dims (with BatchNorm and LeakyReLU).  Two stacked GATConv layers, each with five attention heads, yield context‑aware node embeddings.  A final MLP produces a $(M\!+\!1)$ -way softmax for every UAV, giving a joint action in a single forward pass (see Fig.~\ref{fig:GAT}).  The same encoder is shared by the two critics, which regress separate Q‑values in the TD3 framework.

\subsubsection{Training}
We train for 20,000 episodes on an RTX 3070 Mobile GPU. Mini‑batches are drawn from a replay buffer; the critics follow clipped double‑Q learning while the actor is updated every other critic step, with target‑policy smoothing noise $\mathcal{N}(0,\sigma^2)$ added to the next‑state actions.

\subsection{Deploying the Software}
\subsubsection{AI-based Guidance AI}
The guidance AI model was deployed on an NVIDIA Jetson Orin Nano \cite{NVIDIAJetsonOrinNano2024}, selected for its balance of computational power and energy efficiency, making it suitable for real-time inference on UAVs. Due to the model's lightweight nature, it was implemented directly in Python, as the performance gains from converting the code to C++ were deemed unnecessary for this application.

The system utilised ROS 2 Isaac \cite{NVIDIAIsaacROS2024} to manage communication between the UAV's sensors and control algorithms, facilitating seamless integration between hardware components and the guidance AI model. To interface the AI-generated guidance commands with the PX4 flight controller, the $\mu$XRCE-DDS client \cite{px4_uxrce_dds} was employed with the PX4\_msgs ROS 2 package. This setup enabled the efficient and reliable transmission of control commands from the AI model to the UAV, allowing for the rapid and accurate execution of guidance decisions and avoiding obstacles in real-world environments.

\subsubsection{AI-based Task Allocation}
Task allocation is managed by a central server hosted on a laptop, which communicates with the UAVs over a WiFi connection. Each UAV connects to the server and sends requests for guidance instructions, receiving a path to its next target node from the task allocation model. Due to the fully connected nature of the task allocation graph, the server generates a sequence of nodes to guide the UAV to its target when a direct path between nodes is obstructed. For instance, if a wall blocks the direct route between two nodes, an intermediary node along the feasible path will be included in the sequence to ensure the UAV reaches its destination.

Upon reaching its designated target node, the UAV sends a new request to the server, and the task allocator provides the next target and path. The server maintains a separate graph version for each UAV, dynamically updating these graphs as new targets are assigned. This ensures both UAVs have up-to-date task information, allowing them to coordinate their movements efficiently. Once all nodes have been visited, the server calculates and transmits a return path to the take-off point, facilitating the UAVs' recovery.

\section{Test Results and Analysis}
This section presents the results of the developed UAV navigation, guidance and task allocation systems, divided into three main parts: simulation results, real-world results, and the performance in the Sapience Competition. The simulation results focus on validating the guidance and navigation AI, as well as the performance of the task allocation strategy in a controlled environment. The real-world results evaluate the system's practical performance, including the UAV's ability to navigate around obstacles, follow designated paths, and execute task allocation in real-world conditions. Additionally, the effectiveness of LIDAR-SLAM for height correction and the UAV's capabilities in a mapping task are examined, providing a comprehensive assessment of the system's deployment.

\subsection{Simulation Results}

\subsubsection{Validation of UAV AI-based Guidance}
The training process for the AI-based Guidance generated 260 models that met the criteria for further testing, each having achieved the required number of targeted pose goals. The first stage of validation involved running these models through the training environment and evaluating their performance based on the number of goals achieved. Approximately the top 10\% of models were selected, depending on the distribution of performance. Out of the 260 models, 34 surpassed the chosen threshold of achieving 20 goals across four episodes.

\begin{figure}
    \centering
    \includegraphics[width=0.8\linewidth]{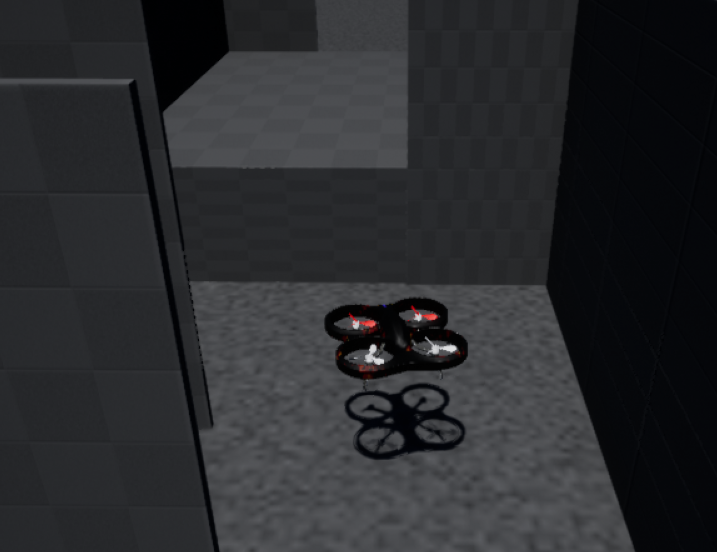}
    \caption{The trained model running through the validation environment}
    \label{fig:droneval}
\end{figure}

These 34 models were then evaluated in a validation simulation designed to more closely resemble the real-world environment in which the UAV would operate. This can be seen in Fig. \ref{fig:droneval}. In this more challenging setting, the best-performing model achieved 304 goals out of a possible 320, with only two crashes recorded during 20 validation runs. The crashes occurred during a particularly difficult manoeuvre, requiring the UAV to drop from an elevated platform and execute a 180-degree turn around a corner.

Notably, the models that reached this second stage of validation emerged from a training window of 77,000 to 170,000 states observed, out of a total of approximately 250,000 states encountered during training. This observation suggests a plateau in training, indicating that further exposure to training data does not necessarily lead to significantly better models. Ultimately, model number 155260 was selected for real-world testing based on its performance in the validation phase.

Along with the APF-rewarded guidance AI, a model was also trained using a more standard distance-based reward, where the reward is given based on how much closer the drone moves to the goal per time-step. Several attempts were made with different hyperparameters and reward weightings, but the training could not produce a model suitable for testing that met the criteria used for the APF-rewarded AI guidance. Due to the time constraints imposed by the competition schedule and the one-week-long training run, the training could not be conducted until convergence.    

\subsubsection{Validation of GAT-based Task Allocation Strategy}
The validation process for the AI-based Task Allocation follows a procedure similar to that used for the AI-based Guidance scheme. During training, models that were able to complete the graph in under six moves were retained for further evaluation. The first stage of validation involved testing these models using the specific graph designed for the real-world scenario, aiming to identify those models that could provide the most efficient solutions.

In the second stage of validation, each selected model was tested on a series of randomly generated graphs. Each model was given five attempts to complete each random graph, and their performance was evaluated based on the total distance travelled across all attempts. The model with the lowest cumulative distance was chosen as the best-performing model, as it demonstrated the most efficient task allocation across diverse scenarios.

The results of this validation gave a model that always achieved success in four steps, meaning no unnecessary double moves. And achieved the lowest average distance travelled when compared to the median. This resulted in a score of -6.0m below the median distance travelled by the selected models.

\color{black}

To contextualise the proposed graph-attention task allocator, we compared its allocation outcome against representative classical baselines using the \emph{same} waypoint-graph cost metric as our method (i.e., edge weights reflect corridor-constrained travel distance, and inter-node costs are shortest-path lengths on the graph). Specifically, we evaluated greedy nearest-task assignment, an auction-style bidding allocator (bids proportional to path cost), and an exact optimal one-step assignment (minimum-cost matching, recomputed after each completed task). On the field scenario used in this paper, all methods produced identical task sequences for both UAVs and therefore identical total travel distance and makespan. This indicates a ceiling effect arising from the small and highly structured competition task set: once true graph distances are used, the optimal allocation becomes largely unambiguous and can be recovered by several standard allocators. We therefore report these results as evidence that the proposed allocator is competitive on the deployed scenario, and we focus the remaining evaluation on deployment-relevant performance factors such as end-to-end autonomy, robustness, and real-world execution constraints.

\begin{table}[t]
\centering
\caption{Task allocation comparison of the GNN with classical methods in the Sapience layout}
\label{tab:task_alloc_baselines}
\begin{tabular}{lcccc}
\textbf{Method} & \textbf{Total distance (m)} & \textbf{Makespan (m)} & \textbf{Steps} \\
Greedy nearest-task & 24.60 & 12.60 & 5 \\
Auction / bidding   & 24.60 & 12.60 & 5  \\
Optimal matching    & 24.60 & 12.60 & 5  \\
\textbf{Ours (GAT-DRL)} & 24.60 & 12.60 & 5  \\
\end{tabular}
\end{table}

\color{black}

\subsection{Testing in a Real-World Environment}
The real-world testing of the UAV systems was conducted in the Autonomous Systems Arena at City, St George’s University of London. For initial tests, the arena was configured as an open space measuring 12 m by 8 m, with large cardboard boxes used as obstacles. These obstacles served to limit potential damage to both the UAV and the testing environment while providing basic challenges for guidance and navigation.

The arena is equipped with Optitrack cameras \cite{optitrack}, which provide precise positional data by tracking reflective markers affixed to the UAV. This setup ensures accurate localisation of the UAV without relying on GNSS, facilitating the evaluation of the adapted LIDAR-SLAM odometry solution adopted in this controlled environment.

Following the initial phase of simple obstacles, a more complex structure was assembled within the arena, featuring a floor plan of 10.8 m by 6 m and walls measuring 2.4 m in height. This structure included a network of 2 m wide corridors and a room measuring 3.6 m by 4.8 m, featuring a 1 m raised floor. This setup was designed to emulate more closely real-world indoor search and rescue challenges. A 2D schematic of the layout is presented in Fig. \ref{fig:layout}. Unlike simulations, where every action is precisely controlled and predictable, real-world environments introduce various sources of uncertainty and unpredictability. 

\begin{figure}
    \centering
    \includegraphics[width=0.8\linewidth]{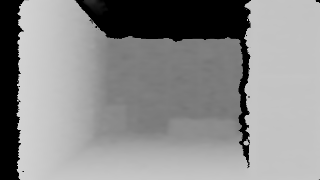}
    \caption{The depth image from the Intel Realsense D455f camera}
    \label{fig:realsense}
\end{figure}

\subsubsection{Real-World Sensor Output}
\begin{figure}
    \centering
    \includegraphics[width=0.8\linewidth]{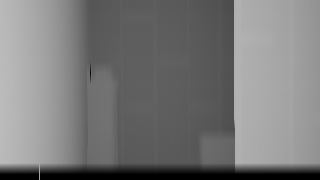}
    \caption{The depth image extrapolated from the Velodyne 16-Line Puck sensor}
    \label{fig:lidardepth}
\end{figure}

\begin{figure}
    \centering
    \includegraphics[width=0.8\linewidth]{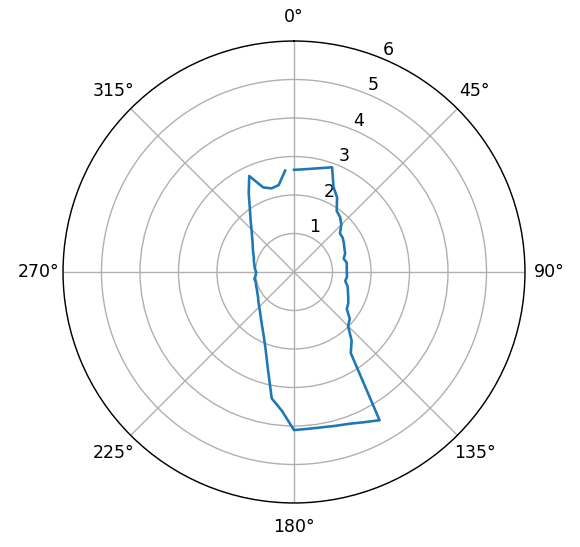}
    \caption{The 2D LIDAR scan used as an input to the guidance AI}
    \label{fig:lidarscan}
\end{figure}

Compared to the simulated sensors, the real-world sensors have to deal with shadows and noise. The Intel Realsense, a stereographic depth camera, has to deal with both issues. The shadows caused by the depth camera are worse at lower ranges, so one way to get a depth image is to interpolate the lidar scan for the area that is required. This gets fewer shadows but must deal with a much lower vertical resolution as this is just 16 lines interpolated to have the same number of vertical pixels as the depth camera. This solution was successfully used for this UAV system. The two outputs are in Fig. \ref{fig:realsense} and Fig. \ref{fig:lidardepth}. Here, the much sharper image from the interpolated lidar point cloud can be seen when compared to the noisy image produced by the Intel Realsense. The interpolated image also lacks the blind spots from having stereoscopic sensors like those used in the Intel Realsense. One problem with using the point cloud is the low vertical field of view; where the Intel Realsense can see the whole wall, the point cloud-derived image can only see the middle section of the wall. A lidar with a larger vertical field of view would fix this issue. Fig. \ref{fig:lidarscan} shows the flat lidar scan used in the AI-based Guidance decision-making, allowing for the 360-degree detection of obstacles.

\subsubsection{Obstacle Avoidance and Path Following}
Initial UAV control tests were conducted in "position mode," where the UAV maintains or moves to a specified position. During these tests, take-off and landing commands were verified, demonstrating stable flight capabilities. For preliminary testing of the AI-based Guidance, the UAV was tasked with reaching specified targets in an unobstructed environment. The measured loop time for the drone’s control function ranged from 0.05 s to 0.07 s, resulting in a control frequency between 20 H and 14 Hz. This revealed overshooting issues, as the UAV struggled to accurately stop within 20 cm of the target due to momentum. To address this, the UAV’s speed was artificially reduced near the target, and its yaw range was expanded, resulting in more reliable target acquisition.

\begin{figure}
    \centering
    \includegraphics[width=0.8\linewidth]{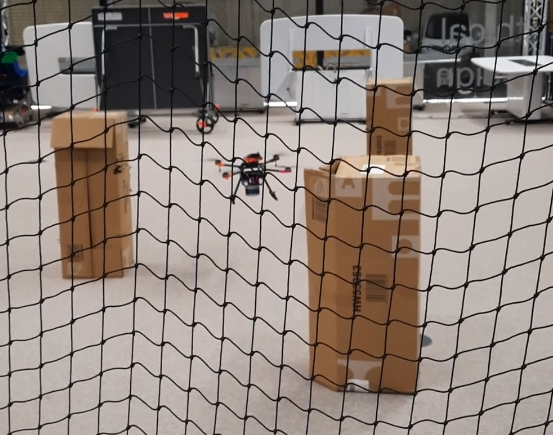}
    \caption{The UAV being tested for obstacle avoidance with the cardboard obstacles}
    \label{fig:tunedrone}
\end{figure}

Next, a single cardboard obstacle was introduced to test basic obstacle avoidance, with the UAV successfully travelling to the target and back. Fig. \ref{fig:tunedrone} shows the UAV flying through the cardboard obstacles. However, with multiple obstacles, the UAV experienced “clipping,” where it came too close to obstacles due to unaccounted momentum. Unlike in simulation, where the UAV could stop and change direction instantly, real-world momentum and inertia needed to be managed. To address this, an emergency avoidance system was implemented. When the 2D LIDAR detected an obstacle within 60 cm, the UAV's velocity was overridden with forces in the opposite direction, defined as follows:

\begin{equation}
v_x = \cos\left(\frac{\pi s}{36}\right) \cdot -0.2,
\end{equation}
\begin{equation}
v_y = \sin\left(\frac{\pi s}{36}\right) \cdot -0.2,
\end{equation}

Where \( s \) represents the section of the LIDAR scan from which the closest response is detected (ranging from 0 to 71), \(v_x\) is the \(x\)-velocity, and \(v_y\) is the \(y\)-velocity. This adjustment significantly improved the UAV’s obstacle avoidance performance.

The testing environment was then expanded with the top half of the structure, including a corridor and a raised section. While the AI-based Guidance was able to change heights as needed, direct control of \(z\)-velocity initially caused oscillation due to overshooting. This was resolved by limiting \(z\)-velocity to \(\pm 0.3 \, \text{m/s}\), resulting in stable vertical movements. Additionally, a condition was added to ensure the UAV travelled fully over the raised ledge before descending, preventing rotor interference with the ledge.

Finally, with the complete structure assembled, the UAV was flown through the environment in various patterns, successfully navigating all sections. These tests confirmed the UAV’s readiness for task allocation and further tests of cooperative functionality and the adapted LIDAR-SLAM odometry.

\begin{figure}
    \centering
    \includegraphics[width=0.8\linewidth]{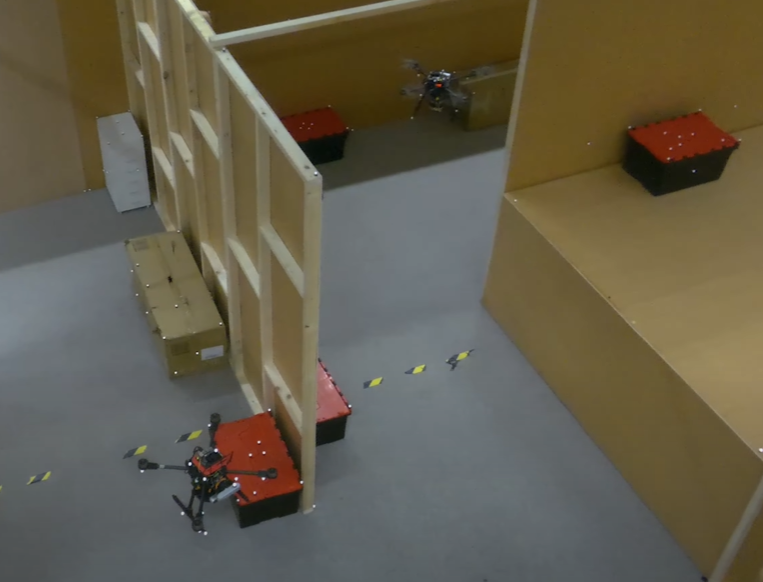}
    \caption{The UAVs moving through the course cooperatively}
    \label{fig:coopdrone}
\end{figure}

\subsubsection{Real-World Task Allocation Performance}
\begin{figure}
    \centering
    \includegraphics[width=0.75\linewidth]{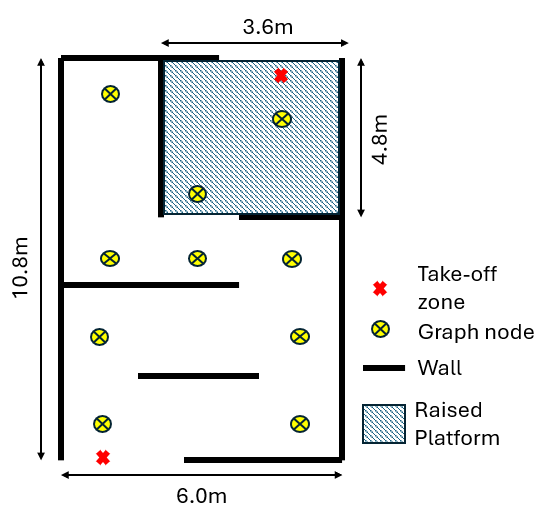}
    \caption{The floor plan for the constructed building with the node locations for the graph}
    \label{fig:layoutnodes}
\end{figure}

To evaluate the task allocation system in a real-world environment, the UAVs were positioned at opposite ends of the building, as shown in Fig. \ref{fig:layoutnodes}. For this task allocation scenario, a 2D floorplan of the building was provided as prior knowledge, allowing nodes to be predefined and the graph to be constructed in advance.

The task allocation server was configured to calculate the shortest viable route between any two nodes that do not have a direct connection. Additionally, the server includes a conflict-detection mechanism to prevent potential path conflicts between UAVs, enhancing safety even though the task allocator is designed to avoid conflicts inherently. 

\begin{figure}
    \centering
    \includegraphics[width=0.75\linewidth]{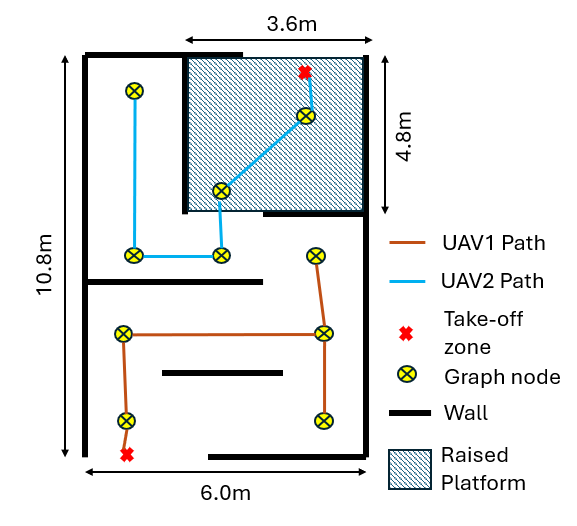}
    \caption{The floor plan for the constructed building with the node locations for the graph}
    \label{fig:layoutpath}
\end{figure}

During testing, the UAVs successfully navigated to their assigned nodes, and the paths they took are illustrated in Fig. \ref{fig:layoutpath}. This setup and testing confirmed the effectiveness of the AI-based task allocation system and the UAVs' ability to execute assigned paths reliably in a real-world environment.

\subsubsection{LIDAR-SLAM Altitude Correction Performance}
To evaluate the UAV's self-localisation performance in a GNSS-denied environment, the adapted LIDAR-SLAM solution was tested within the building. Since GNSS positioning is unavailable in the indoor testing arena, an alternative ground truth was established using the OptiTrack system. This system provides positional tracking by using multiple cameras to monitor reflective markers arranged in unique patterns on the UAV.

The high walls of the testing structure, however, can occasionally obstruct the OptiTrack system, degrading its tracking accuracy. This occlusion can lead to small positional errors and, in cases of complete occlusion, a temporary loss of positioning data.

\begin{figure}[ht]
    \centering
    \includegraphics[width=1.0\linewidth]{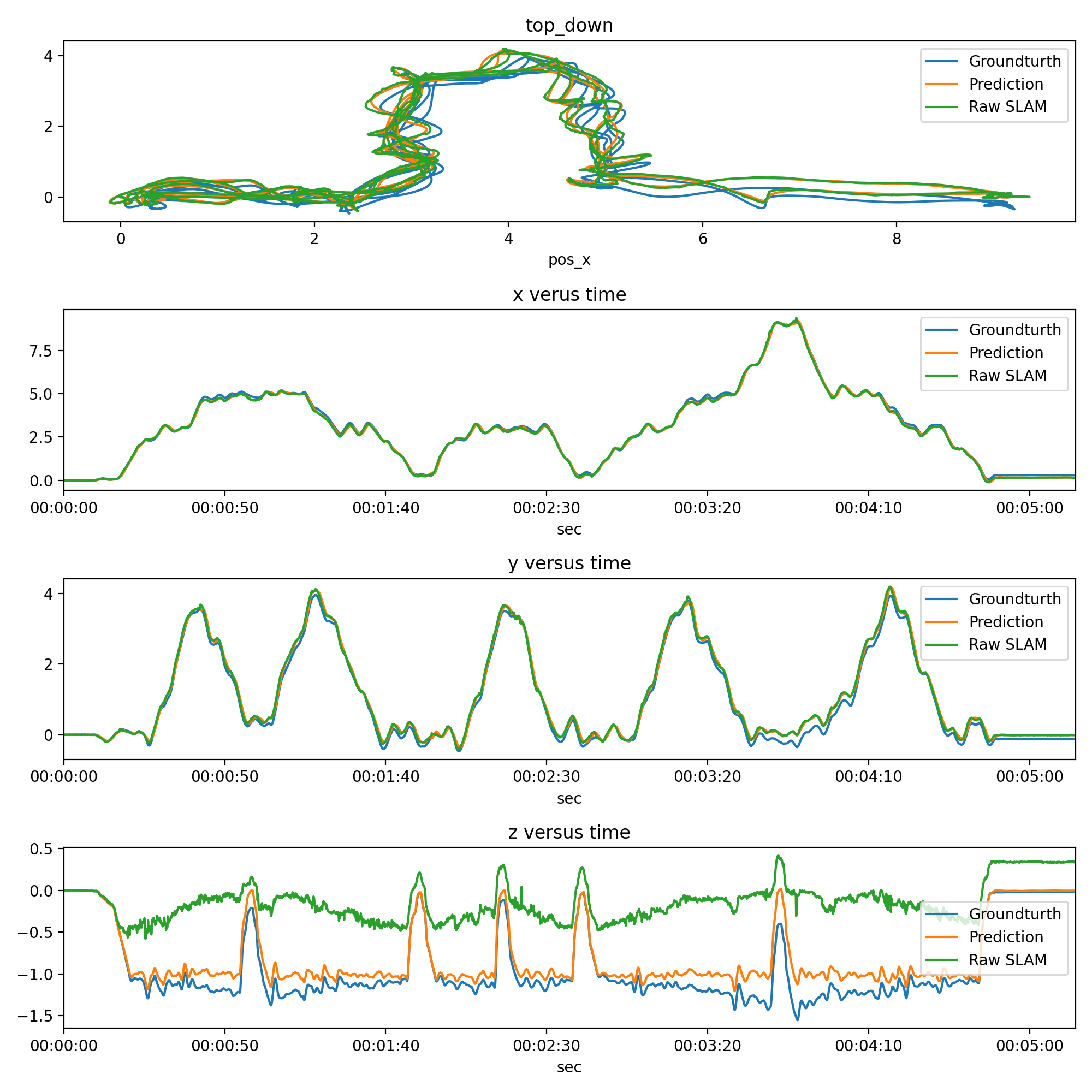}
    \caption{Height corrected LIDAR-SLAM comparison with the ground truth and the raw LIDAR-SLAM}
    \label{fig:lidarresults}
\end{figure}

Initial testing showed that the LIDAR-SLAM solution performed accurately in the \(x\) and \(y\) directions but struggled with \(z\)-axis localisation. To address this, the methodology described in the LIDAR-SLAM Altitude Fix portion of the Methodology (VI.B) was implemented to correct the \(z\) positioning. In Fig. \ref{fig:lidarresults}, the positional trace for each axis is shown over time. The raw LIDAR-SLAM data aligns closely with the ground truth in \(x\) and \(y\) positions, demonstrating minimal deviation. However, significant discrepancies are evident in the \(z\)-axis positioning; the raw LIDAR-SLAM data fail to maintain an accurate \(z\) position soon after take-off and do not regain reliable positioning throughout the test.

With the applied corrections, the predicted \(z\)-position from the modified LIDAR-SLAM solution shows a marked improvement, achieving better alignment with the ground truth. Notably, the only visible deviations occur when the ground truth itself displays errors in tracking, which is attributed to temporary occlusions within the OptiTrack system. Evidence of this error can be seen when the UAV returns to land at consistent heights, indicating reliable \(z\)-positioning despite intermittent inaccuracies in the ground truth data.

In Fig. \ref{fig:3dgraph}, the UAV's height transition from the upper platform to floor level and back again is accurately maintained. The UAV successfully adjusts its height relative to the world coordinate system, demonstrating effective height control. This result indicates that the comparison between IMU-derived velocity and calculated velocity enabled the appropriate adjustment of the height offset, allowing for stable vertical positioning during transitions.

\subsection{The Sapience Autonomous Cooperative Drone Competition}
The Sapience competition took place over a week in August 2024 at City, St George’s University of London, with four teams from NATO and NATO partner countries participating. The competition consisted of three tasks designed to simulate a search and rescue operation within a specially constructed indoor arena. Emphasis was placed on the speed of task completion to encourage cooperative use of the two UAVs made available for use in this competition. Teams were additionally evaluated on the innovation in their solutions and the quality of their system outputs.

The first task required each team to produce a legible 3D arena map quickly. In the second task, teams were tasked with detecting and localising three mannequins and five boxes within the arena, with accuracy in detection and localisation serving as key evaluation criteria. The final task involved completing eight deliveries to the mannequins, with the last two deliveries requiring synchronisation between the UAVs. Teams were assessed on the number of successful deliveries and the speed of completion within the allotted time. These tasks simulate essential search and rescue activities, such as those required in the aftermath of an earthquake.

This paper focuses exclusively on the first task and third tasks, as they show the ability of the UAVs to perform autonomous cooperative flight better than the second task. The second task only entailed a short pre-planned flight to collect data for the detection and localisation systems. As these systems are not covered in this paper, this task will be skipped.

\subsubsection{Review of System Performance in 3D Mapping Task}

The first task, generating a 3D map of the environment in the shortest time possible, provides an opportunity to evaluate all system components. This task demonstrates the UAVs' ability to efficiently and safely navigate the environment cooperatively, the task allocator's capacity to guide the UAVs to achieve full coverage, and the GNSS-denied capabilities of the system. Additionally, the UAVs' capability to handle height variations necessary for complete mapping was assessed.

\begin{figure}[t]
    \centering
    \includegraphics[width=1.0\linewidth]{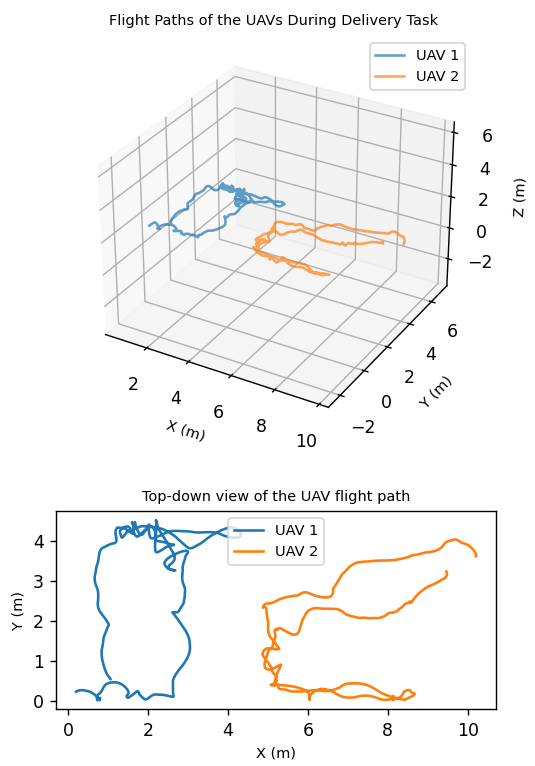}
    \caption{The path of the two UAVs through the building whilst building the map}
    \label{fig:3dgraph}
\end{figure}

In Fig. \ref{fig:3dgraph}, the positions of the UAVs during the mapping task are shown. Position data was derived from the corrected LIDAR-SLAM, with coordinates relative to the origin point at \((0,0,0)\), located at the lower-level entrance take-off position shown at the bottom of Fig. \ref{fig:layoutpath}. The second take-off position is on the raised platform around \((10.5, 3.2, 1.0)\).

The first graph illustrates the 3D paths both UAVs take throughout the building. UAV 1 navigates around the lower section of the building, effectively completing a circuit around the free-standing wall. UAV 2, starting from the raised platform, descends to floor level, explores the long corridor, and then returns to its starting area.

The second graph provides a top-down view of the UAVs’ flight paths. UAV1 tends to struggle at the junction of the arena as it has to deal with many obstacles from various angles. This often required it to try multiple efforts to get through the gap between the lower inner wall and the eastern outer wall. The flight is not as smooth as a preplanned route, but this method shows an AI-based Guidance ability to control a UAV's navigation in real time in a realistic search and rescue situation.

\begin{figure}[t]
    \centering
    \includegraphics[width=1.0\linewidth]{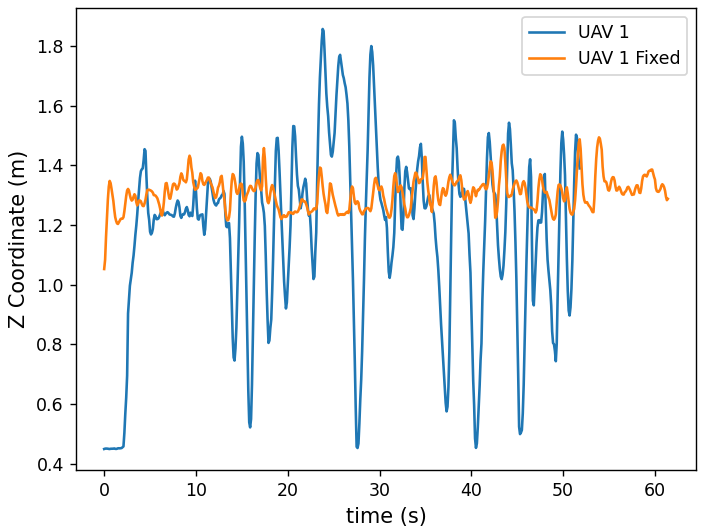}
    \caption{The drone's height during a flight compared to a flight before the overheating was fixed}
    \label{fig:height_fix}
\end{figure}

In Fig. \ref{fig:height_fix}, a comparison of two flights of UAV 1 reveals significant differences in performance. The first flight occurred before the competition and was marked by overheating issues in the electronic speed controllers (ESCs). This resulted in the UAV exhibiting erratic behaviour while completing its tasks. In contrast, the second flight we did for the competition, during which the ESCs were relocated to be secured to the arms of the UAV, illustrates a notable improvement in height control following this modification to the ESC configuration. UAV 2, which retained the older ESC layout, did not experience similar problems, indicating a potential assembly fault in UAV 1. UAV 2 was also upgraded to the new ESC layout to mitigate future complications. This adjustment enhanced the UAV's repairability by disconnecting the ESCs without disassembling the main body.

\begin{figure}
    \centering
    \includegraphics[width=1.0\linewidth]{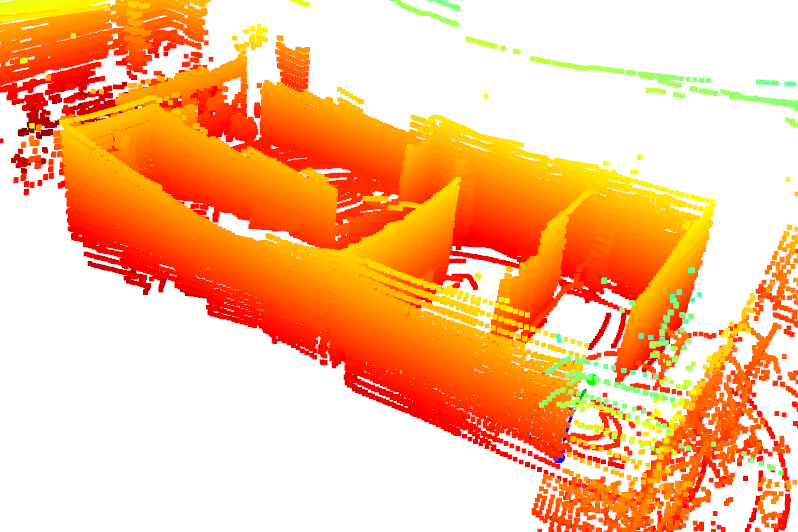}
    \caption{The 3D point cloud gathered from the cooperative UAV flight}
    \label{fig:pointcloud}
\end{figure}

The resulting 3D point cloud map, shown in Fig. \ref{fig:pointcloud}, illustrates the completed map. The second UAV was provided with an approximate relative position for its odometry to align the two point clouds from each UAV. Final alignment was achieved through registration to produce a coherent map. The mapping task was completed in 134 seconds, whereas the same task using a single UAV took 287 seconds, a time reduction of 53.3\%.

Since this task focused on mapping speed, the point cloud density is relatively low. However, this density could be increased by reducing the UAVs' top speed, increasing the number of nodes, or adjusting their height variations. These adjustments would allow for a denser, more detailed map if necessary.

\subsubsection{Delivery Task System Performance}

The delivery task required the UAVs to complete eight deliveries to three designated points within the arena. A delivery was successful when a UAV flew to a target point, landed for a predefined duration, and then returned to its starting position. Additionally, the final two deliveries were required to be completed simultaneously, demonstrating the ability to coordinate the cooperative actions of both UAVs effectively.

\begin{figure}
    \centering
    \includegraphics[width=1.0\linewidth]{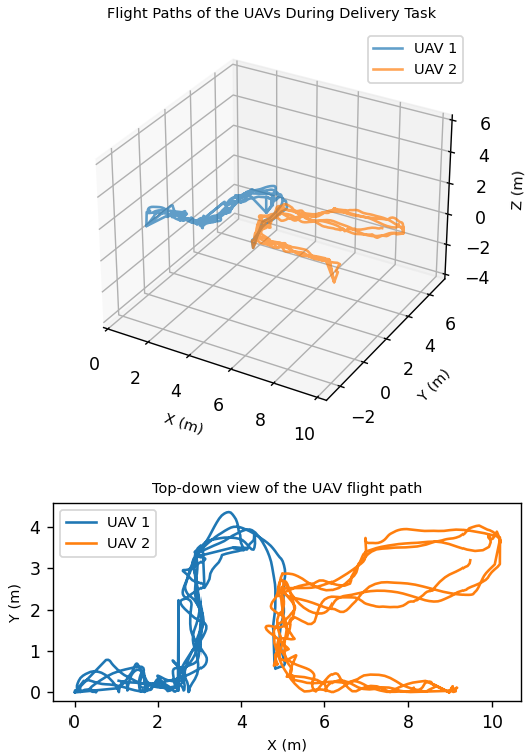}
    \caption{The path of the two UAVs through the building whilst doing the delivery task}
    \label{fig:delivery_path}
\end{figure}

Fig. \ref{fig:delivery_path} illustrates the paths taken by the UAVs during the delivery task. To prevent potential conflicts at the first delivery point, UAV 1 was dispatched first to complete its delivery, followed by UAV 2 after UAV 1 had cleared the point. Once UAV 2 completed its first delivery, UAV 1 resumed its remaining deliveries. To improve synchronisation, UAV 1 was delayed by 10 seconds at each take-off, as UAV 2 had longer distances to travel between delivery points.

For the final synchronised deliveries, the UAVs employed a coordination protocol. The first UAV to arrive at its delivery point hovered above it and sent a "ready" signal to the server. The server held the UAV in a hover state until it received a matching "ready" signal from the second UAV. Once both UAVs signalled readiness, the server issued a simultaneous landing command, ensuring synchronised landings.

\begin{figure*}
    \centering
    \includegraphics[width=0.9\linewidth]{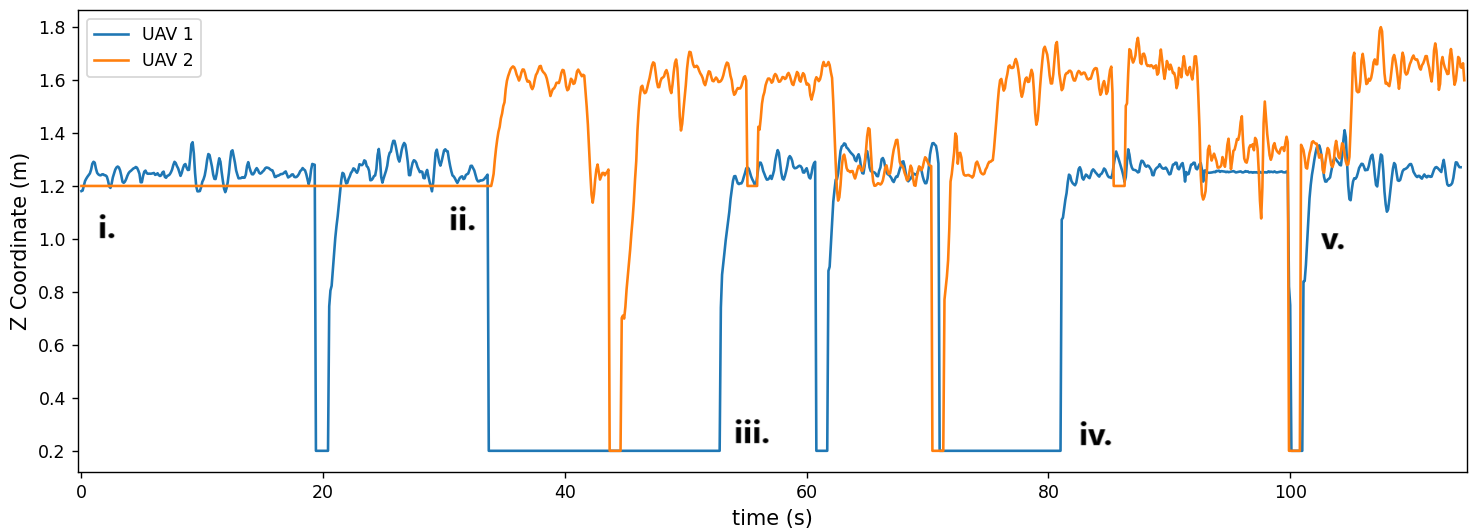}
    \caption{The height (z) against time to show the deliveries and the final synced delivery. (i) UAV 2 does not take off until UAV 1 has completed its first delivery. (ii) Once UAV 1 is landed, UAV 2 takes off to complete its task. (iii) As UAV 2 makes the last conflicted delivery, UAV 1 starts again but is delayed by 10s to better sync the deliveries at the end. (iv) The UAV 1 delayed take-off when at base. (v) The sync delivery takes place once the UAVs detect they have both reached the delivery point.}
    \label{fig:height_delivery}
\end{figure*}

Fig. \ref{fig:height_delivery} shows the timing of the deliveries, highlighting the UAVs’ landing altitudes of 0.2 m on the floor and 1.2 m on the raised platform. Despite the delayed take-offs for UAV 1, a 10-second loiter period was still necessary to ensure synchronised landings. Compared to the mapping task, the UAVs' speeds were increased to a maximum of 1 m/s, prioritising faster delivery times as it can be critical to the mission's success in a real-world search and rescue scenario.

\begin{figure}
    \centering
    \includegraphics[width=1.0\linewidth]{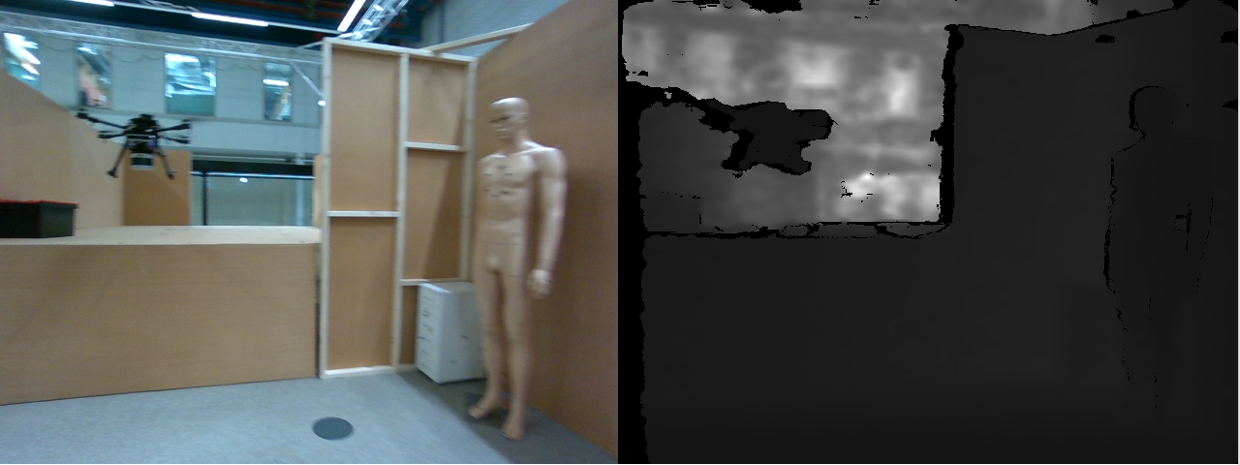}
    \caption{A head-to-head view of a drone moving from the raised section to the lower section}
    \label{fig:doubleimg2}
\end{figure}

Fig. \ref{fig:doubleimg2} shows the outputs from the detection cameras during the delivery task, capturing both UAVs en route to their respective delivery points.

\color{black}
\subsection{Demonstration on an Alternative Layout}
To provide additional evidence that the proposed cooperative framework is not specific to a single arena geometry, we include a brief demonstration on a secondary indoor layout used for system checks and demonstrations. This layout is intentionally simpler than the Sapience competition arena due to the complexities of erecting the Sapience arena and is therefore not presented as a stronger benchmark. Rather, it serves as a compact sanity check that the same end-to-end pipeline (waypoint graph construction, graph-based task allocation, and learned guidance execution) can be applied without altering the controller structure or task allocation logic.

Figure~\ref{fig:alt_layout_map} shows the alternative layout and the set of task waypoints. The waypoint graph is constructed in the same manner as in the competition setting, and allocator decisions are computed using a GNN to determine the most efficient way to visit all nodes and complete the cooperative point cloud. The arena consists of two drones, one at each end, separated by a T-shaped obstacle in the middle with three mannequins arranged around it. The drones are also commanded to land at the opposite take-off site to reduce the time required to complete the task.

Table~\ref{tab:task_alloc_demo} compares the proposed GAT-DRL allocator against representative classical methods on the demonstration layout, using shortest-path distances on the same waypoint graph as the cost metric. The auction/bidding and optimal matching baselines yield identical performance in this instance, with a total travel distance of \(26.30\,\mathrm{m}\) and a makespan of \(14.88\,\mathrm{m}\) (where makespan denotes the maximum distance assigned to any single UAV). In contrast, the proposed allocator reduces the total travel distance to \(21.19\,\mathrm{m}\) and the makespan to \(11.94\,\mathrm{m}\), corresponding to reductions of \(19.4\%\) and \(19.8\%\), respectively. The proposed method also completes the allocation in fewer steps between nodes (4 versus 7), indicating a more stable and efficient division of work between the two UAVs on this layout.

\begin{figure}[t]
    \centering
    \includegraphics[width=\linewidth]{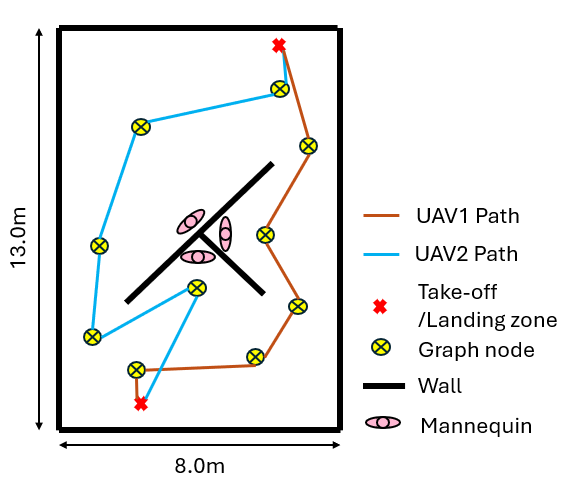}
    \caption{Waypoint-graph visualisation of the Demo layout, showing the routes executed by each UAV.}
    \label{fig:alt_layout_routes}
\end{figure}

\begin{figure}[t]
    \centering
    \includegraphics[width=\linewidth]{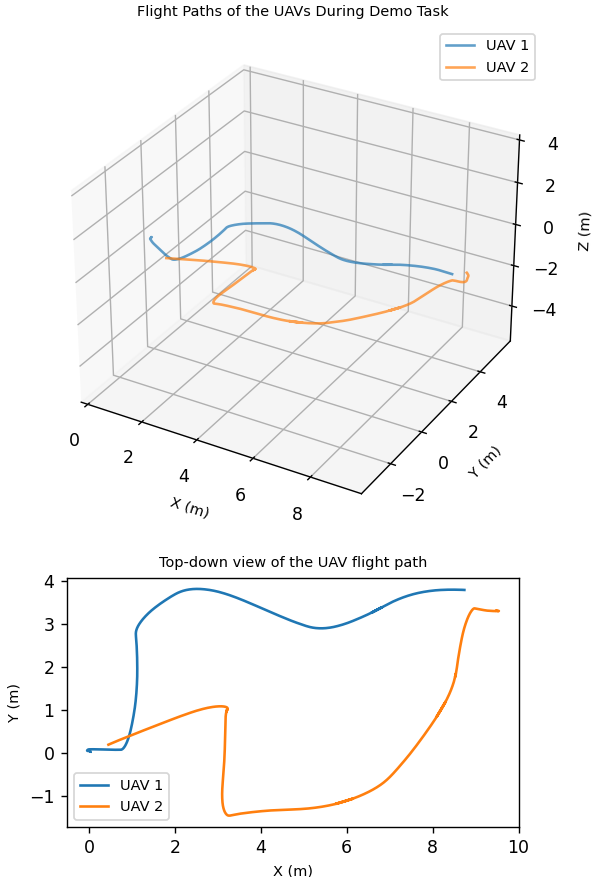}
    \caption{The routes of the two UAVs when completing the Demo task.}
    \label{fig:alt_layout_map}
\end{figure}

Figure~\ref{fig:alt_layout_routes} summarises the resulting routes from the odometry as measured by the UAVs. Although this layout is simpler, the result demonstrates that the allocator and guidance policy remain coherent under a change in arena geometry, producing consistent task separations between agents and feasible collision-free routes.

\begin{figure}[t]
    \centering
    \includegraphics[width=\linewidth]{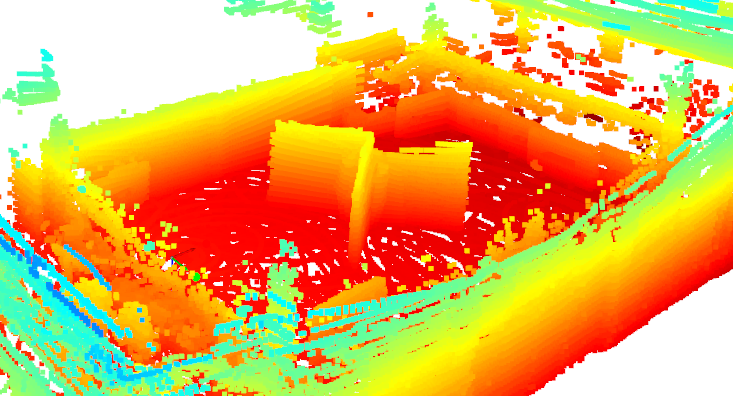}
    \caption{The combined 3D point cloud collected from both UAVs as they have completed their allotted tasks.}
    \label{fig:alt_layout_pointcloud}
\end{figure}

To complement the top-down layout depiction, Fig.~\ref{fig:alt_layout_pointcloud} shows the aggregated 3D point cloud collected during the Demo-layout flight. 

\begin{table}[t]
\centering
\caption{Task allocation comparison of the GNN with classical methods in the Demo layout}
\label{tab:task_alloc_demo}
\begin{tabular}{lcccc}
\textbf{Method} & \textbf{Total distance (m)} & \textbf{Makespan (m)} & \textbf{Steps} \\
Greedy nearest-task & 29.26 & 14.64 & 7 \\
Auction/bidding & 26.30 & 14.88 & 7  \\
Optimal matching    & 26.30 & 14.88 & 7  \\
\textbf{Ours (GAT-DRL)} & 21.19 & 11.94 & 4  \\
\end{tabular}
\end{table}

\color{black}

\section{Additional Lessons Learned from the Deployment of the UAVs}

The development and deployment of the UAV system for the SAPIENCE competition revealed several critical areas that required adjustment to ensure improved operation. Key challenges included optimising AI-based guidance, implementing the artificial potential field (APF) to improve collision avoidance, and limiting the training environment.

\subsection{Tuning the AI-based Guidance}

The initial direct embedding of the AI-based guidance highlights the need for tuning specific control outputs, especially in the vertical (z-axis) velocity. Without adjustment, the model's output was too aggressive for the flight controller, causing oscillations around the target altitude. Limiting the z-axis velocity command to ± 0.3 m/s
resolved this issue, allowing the UAV to maintain smoother and more stable flight near target altitudes. Additionally, adjustments in lateral (x-axis) velocity and yaw speed near targets improve the UAV’s accuracy in reaching and stabilising at target points, increasing success rates in approaching waypoints without overshooting.

\subsection{Implementation of the Artificial Potential Field (APF)}

To ensure safe navigation, an APF layer was added to the system as an additional safeguard against collisions. While the guidance AI was trained with an obstacle avoidance objective, real-world dynamics introduced momentum and other factors that required more robust safety measures than initially provided by the reward structure. The APF’s incorporation points to the potential need for refining the reward structure during training so that avoidance behaviours are better aligned with the complexities of real-world flight dynamics.

\subsection{Limitations of the Training Environment}

A significant insight from this deployment was the importance of the quality of the simulation environment. The training and validation environments used in AirSim were simple and often failed to capture the complexities encountered in the real world. Real-world scenes with larger open spaces, varied materials, and complex textures—such as glass and netting—exposed gaps in the depth camera’s data, presenting challenges not accounted for in the training phase. Due to the requirements of AirSim, configuring highly detailed environments with realistic depth textures proved challenging, as each texture must be specifically adapted for depth imaging to function correctly.

\section{Future Work}
The three-stage NATO Sapience programme provides a natural roadmap for extending our framework beyond the present indoor demonstration.

\subsection{Outdoor and Mixed Indoor--Outdoor SAR}
The second competition will be held entirely outdoors, followed by a combined indoor/outdoor finale.  Operating without Global Navigation Satellite Systems (GNSS) in feature--sparse exteriors demands: (i)~\textbf{fully on--board autonomy}---communication with the base station will be severed, so all high‐level decisions must arise from inter‐UAV links; (ii)~\textbf{richer odometry}, e.g.\ visual--inertial--LiDAR fusion to maintain drift below~$0.5\,\%$ of distance travelled; and (iii)~\textbf{ruggedised airframes} capable of lifting heavier battery and sensor payloads (~7kg MTOW).

\subsection{Multi--Agent Deep RL Guidance}
The present work trains independent TD3 agents that share state vectors at inference time.  Future research will explore Multi‐Agent DRL (MADRL) objectives that explicitly model pairwise collision avoidance and cooperative coverage through joint value functions or centralised critics. 

\subsection{Decentralised Task Allocation}
With the base station removed, our Graph Attention Network must execute on board.  Two deployment topologies will be studied: a master–slave scheme in which a single UAV performs inference and broadcasts assignments, and a fully peer‐to‐peer consensus that amortises computation across the fleet and eliminates a single point of failure.

\subsection{Collaborative Perception and Mapping}
We will extend the depth–IMU fusion that eliminates indoor LiDAR Z‐drift to a multi‐agent odometry pipeline.  Sharing pose graphs and loop closures over a low‐bandwidth mesh could reduce accumulated vertical error by an order of magnitude, enabling altitude‐critical tasks such as window ingress.

These directions will culminate in a robust, decentralised multi-UAV SAR stack capable of seamless indoor–outdoor transition and long-range operation without external infrastructure.

Moving past the competition setting, several research directions will mature our framework into a general‑purpose indoor autonomy stack.  First, we will replace the current waypoint-based navigation with goal-conditioned multi-agent DRL policies that learn cooperative, obstacle-aware trajectories directly from graph-structured sensor observations.  This shift will allow each UAV to reason jointly about its neighbours and the environment in a unified latent space, eliminating the need for predefined intermediate goals.

Safety at close quarters will be bolstered through the real‑time exchange of lightweight perceptual cues—such as truncated LiDAR point clouds and depth key‑frames—so that each agent can augment its local occupancy map with the perspectives of its teammates.  Shared perception is expected to reduce occlusion‑induced collisions and enable tighter formations in narrow corridors.

Accurate state estimation remains a bottleneck for large‑scale indoor operations.  Building on our depth–IMU fusion, we will extend the filter to a fleet-level optimisation that incorporates inter-agent loop closures and ultra-wide-band ranging, yielding a globally consistent altitude and eliminating residual LiDAR-SLAM Z-drift.

Collectively, these avenues will transform the current prototype into a robust, scalable platform for collaborative UAV autonomy in cluttered, GNSS-denied environments.

\section{Conclusion}

This work presents an extensive real-world demonstration of a cooperative, DRL-guided UAV system operating in GNSS-denied indoor environments. We integrated LiDAR-SLAM for on-board localisation and mapping, a velocity-level DRL guidance policy shaped by an APF reward, and a centralised GAT-based allocator for multi-UAV coordination—then validated the complete stack under competition pressure, rather than in simulation alone.

The NATO SAPIENCE Autonomous Cooperative Drone Competition provided a realistic proxy for SAR deployments. Across mapping, object detection, and synchronised delivery tasks, our system showed it can: (i) rapidly produce actionable 3‑D maps of complex structures, (ii) coordinate time‑critical, multi‑agent behaviours, and (iii) sustain reliable autonomy amidst clutter, occlusions, and tight passages. These trials exposed—and allowed us to solve—practical hurdles that seldom appear in simulation studies, including emergency obstacle-avoidance fallbacks, thermal management, and altitude drift suppression via depth–IMU fusion.

By moving beyond constrained 2D controllers, small arenas, or purely waypoint-level actions, we demonstrate that DRL policies can scale to full-scale airframes, richer state-action spaces, and prolonged indoor missions. The resulting framework provides a concrete blueprint for deploying autonomous, cooperative UAV teams in life-saving SAR scenarios and other GNSS-denied operations.

Overall, our findings underscore a central message: closing the sim-to-real gap requires end-to-end field experimentation. The lessons, architectures, and reward designs reported here provide a tested foundation on which future work—both within and beyond SAPIENCE—can build toward faster, safer, and more effective collaborative aerial robotics.

\section*{Acknowledgment}
We would like to thank NATO for helping to organise the SAPIENCE competition, the various staff at City St George's University of London for their help in setting it up, and the other competition teams from Delft University of Technology, the University of Alabama in Huntsville, and the University of Klagenfurt for their excellent work and help in making it a success. I would finally like to thank the other members of our RAMI research group for their patience and help while organising the competition.

\bibliographystyle{plain}
\bibliography{refs}

\end{document}